\RequirePackage{fix-cm}
\documentclass[twocolumn]{galbot}

\usepackage{amsmath,amssymb,array,tabularx,colortbl,xspace,xurl}
\makeatletter
\@ifundefined{onedot}{%
  \def\@onedot{\ifx\@let@token.\else.\null\fi\xspace}%
  \DeclareRobustCommand\onedot{\futurelet\@let@token\@onedot}}{}
\makeatother

\providecommand{\ie}{\emph{i.e}\onedot}

\providecommand{\method}{ALV\xspace}
\providecommand{\alvl}{ALV-L\xspace}   
\providecommand{\alvls}{ALV-L-S\xspace} 
\providecommand{\best}[1]{\textbf{#1}}
\providecommand{\second}[1]{\underline{#1}}

\renewcommand{\best}[1]{{\rmfamily\bfseries #1}}
\newsavebox{\alvfitbox}
\newcommand{\fitwidth}[1]{\sbox{\alvfitbox}{#1}%
  \ifdim\wd\alvfitbox>\linewidth\resizebox{\linewidth}{!}{\usebox{\alvfitbox}}\else\usebox{\alvfitbox}\fi}
\definecolor{alvA}{HTML}{C25500}   
\definecolor{alvL}{HTML}{0072B2}   
\definecolor{alvV}{HTML}{00846B}   
\definecolor{alvGrey}{HTML}{6B7280} 
\definecolor{alvBg}{HTML}{F3F4F6}   
\colorlet{alvOurs}{galbot!10} 

\crefname{section}{Sec.}{Secs.}
\Crefname{section}{Section}{Sections}
\crefname{table}{Tab.}{Tabs.}
\Crefname{table}{Table}{Tables}
\crefname{figure}{Fig.}{Figs.}
\Crefname{figure}{Figure}{Figures}
\crefname{equation}{Eq.}{Eqs.}
\Crefname{equation}{Equation}{Equations}
\titleformat*{\subsection}{\large\sffamily\bfseries\raggedright}
\usepackage{needspace}

\usepackage{balance}
\makeatletter
\AtBeginEnvironment{thebibliography}{%
  \pretocmd{\bibitem}{%
    \ifnum\value{page}=\getpagerefnumber{alv:bib-end}\relax
      \if@firstcolumn\balance\fi
    \fi}{}{}}
\makeatother
\AtEndEnvironment{thebibliography}{\label{alv:bib-end}}
\AddToHook{cmd/appendix/before}{\nobalance}

\title{ALV: Advancing Whole-Body Intelligence in Reverse, from Action to Language to Vision}
\hypersetup{pdftitle={ALV: Advancing Whole-Body Intelligence in Reverse, from Action to Language to Vision}}
\author[1, 3, *]{Zhuo Cao}
\author[1, 3, *]{Yingsheng Zhang}
\author[1, 3, *]{Zekun Qi}
\author[1, 3]{Xuchuan Chen}
\author[1, 3]{Zekai Li}
\author[2, 3]{Dairu Liu}
\author[1, 3]{Ruixi Yu}
\author[2, 3]{Luyang Niu}
\author[2, 3, \dagger]{He Wang}
\author[1, 3, \dagger]{Li Yi}

\affiliation[1]{Tsinghua University}
\affiliation[2]{Peking University}
\affiliation[3]{Galbot}

\contribution[*]{Equal contribution, order decided randomly}
\contribution[\dagger]{Corresponding author}

\date{\today} 
\page{\url{https://galaxygeneralrobotics.github.io/alv}}
\abstract{
Vision-language-action and world-action models have advanced robotic manipulation by learning from large-scale demonstrations. Yet extending this recipe to humanoids is challenging: whole-body demonstrations are scarce, and imperfect motion can compromise balance. We argue that whole-body intelligence should instead be built from the body up, establishing reliable motor capabilities before introducing language and vision. We present ALV, a progressive framework that learns Action, Language, and Vision in this order. First, we construct an execution-grounded action space from tracker-verified motion. We then train a language-conditioned flow policy that generates and continually reconditions whole-body motion from instructions and proprioception, enabling interactive control as intent changes. Finally, we introduce visual conditioning and post-train the generative backbone on teleoperated demonstrations. Experiments show that ALV generates realistic motions with strong execution robustness across trackers, perturbations, and long horizons, while significantly improving online instruction switching. Furthermore, the learned motion prior enhances subsequent VLA training, achieving first-class performance among humanoid models. ALV offers a bottom-up path toward humanoid whole-body intelligence, where higher-level capabilities build upon reliable motion.
}

\begin{document}
\pagestyle{plain}
\twocolumn[{%
  \begin{minipage}{\textwidth}
    \mymaketitle
    \par\vspace*{1.25\baselineskip}
  \end{minipage}%
}]
\raggedbottom
\begin{figure*}[!t]
  \centering
  \includegraphics[width=\textwidth]{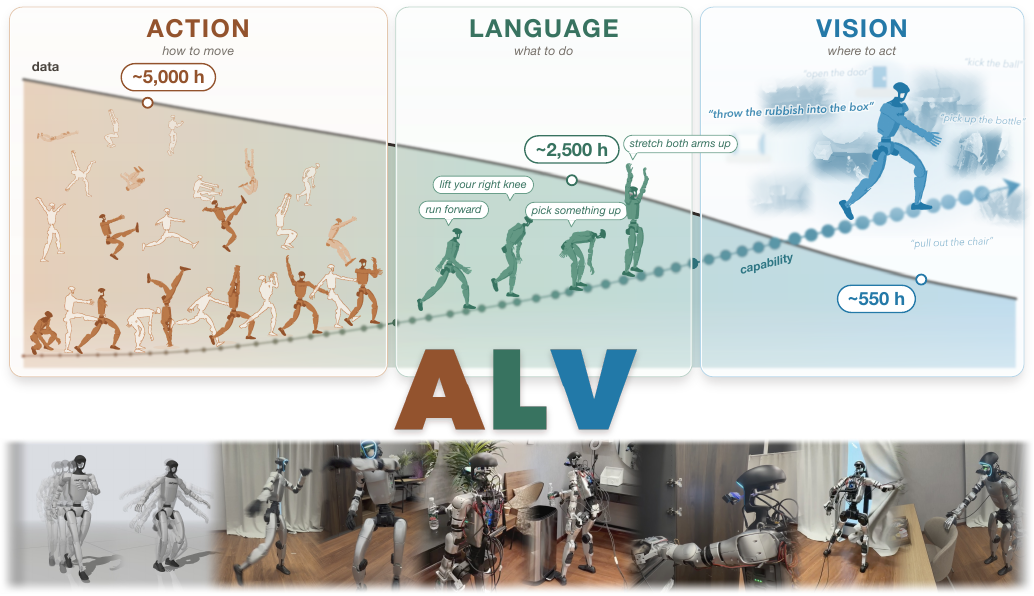}
  \caption{\textbf{Whole-body intelligence from the body up.} \method progressively learns Action, Language, and Vision: how to move, what to do, and where to act. The overview connects each stage's data and capabilities, with examples of language-driven whole-body motion and vision-conditioned action sequences. The data volumes are estimates of publicly available data in the community.}
  \label{fig:teaser}
\end{figure*}

\section{Introduction}
\label{sec:intro}

Robotic manipulation has advanced rapidly in recent years, driven by vision-language-action (VLA) models and world-action models (WAMs)~\cite{brohan2023rt2,kim2024openvla,qin2026wbwam}. A common recipe is to leverage the rich visual and semantic priors of pretrained vision-language or video models and learn action generation from large collections of demonstrations~\cite{brohan2023rt2,kim2024openvla,black2025pi0,bjorck2025gr00t}. Teleoperation, portable data-collection interfaces, and egocentric demonstrations have made it increasingly practical to collect paired observations, instructions, and robot actions~\cite{zhao2023act,chi2024umi,fu2024humanplus,ze2025twist}. Together, these advances have enabled increasingly general end-to-end manipulation policies.

Extending this recipe to humanoid whole-body intelligence, however, presents two fundamental challenges.

\textbf{First, whole-body demonstrations are far harder to collect.} Unlike a fixed-base manipulator, a humanoid must coordinate its arms, legs, and body while maintaining balance and managing contacts. General-purpose whole-body controllers have only recently become capable of tracking diverse human motions~\cite{qi2026humanoidgpt,liao2026beyondmimic,luo2026sonic}, and collecting demonstrations often requires specialized motion-capture equipment. Even with such systems, imperfect tracking limits the range of behaviors that can be reliably demonstrated. Consequently, diverse whole-body demonstrations with synchronized vision, language, and actions remain scarce.

\textbf{Second, whole-body control is far less forgiving of imperfect motion.} A fixed-base arm can often pause, hesitate, or correct its trajectory without compromising stability. A humanoid executing dynamic movements cannot afford the same freedom: small errors in balance, foot contact, or body motion can accumulate into instability or falls. A motion sequence that looks plausible kinematically may therefore be impossible to execute reliably~\cite{araujo2026gmr,he2025asap}. For humanoids, motion quality is not merely a matter of smoothness or realism; it is a prerequisite for successful behavior.

These challenges suggest that whole-body intelligence should be built \emph{from the body up, rather than from vision down}. Instead of asking a visually pretrained model to learn both what to do and how to move from scarce demonstrations, we argue for establishing reliable motor capabilities before introducing higher-level conditioning. Importantly, the available data naturally supports this order. Human motion is abundant in motion-capture archives, generated sequences, and curated motion datasets~\cite{mahmood2019amass,fan2025gotozero,hymotion2025,huang2026omg}. Motion paired with language is less abundant, while demonstrations that additionally include egocentric vision are the scarcest. This motivates a progressive learning strategy: first learn \emph{how to move}, then \emph{what movements language describes}, and finally \emph{how to adapt those movements to the visual world}.

\begin{figure*}[t]
  \centering
  \vspace{5pt}
  \includegraphics[width=\textwidth]{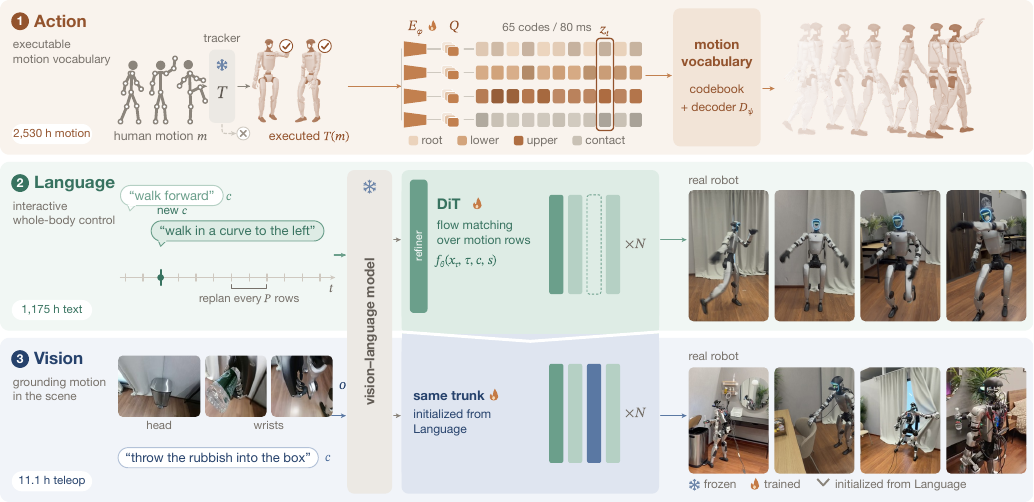}
  \caption{\textbf{ALV builds whole-body intelligence from the body up.} (1) \textbf{Action:} tracker-verified trajectories train a four-stream residual-VQ motion vocabulary and decoder. (2) \textbf{Language:} a flow-matching policy generates and revises motion from instructions, robot state, and committed motion. Bounded-lookahead decoding and rolling replanning support continuous execution as intent changes. (3) \textbf{Vision:} a pretrained Qwen3-VL-based backbone provides visual context to the transferred language-stage generative trunk, which is adapted to predict native robot actions. The legend distinguishes frozen, trained, and language-initialized components.}
  \label{fig:method}
\end{figure*}

Following this principle, we introduce \method, a bottom-up framework that builds whole-body intelligence through three successive stages: \emph{Action, Language, and Vision} (\cref{fig:teaser}). Each stage builds upon the representations learned in the previous one, progressively introducing richer conditioning without learning motor capabilities from scratch.

In the \textbf{Action} stage, we first construct a reliable motion vocabulary. Rather than directly modeling human motion, which may be physically infeasible after retargeting, we execute candidate motions through a whole-body tracker and retain only faithfully tracked trajectories. A four-stream residual vector-quantized tokenizer then learns a compact representation of these executed motions, providing an action space grounded in what the robot can actually perform. In the \textbf{Language} stage, we train a flow-matching diffusion transformer to generate motion in this space from language instructions, acquiring a broad repertoire of language-conditioned whole-body behaviors without requiring visual demonstrations. Because the policy is conditioned on proprioception and on motion already committed for execution, transition-aware post-training lets the same network revise its future motion when the instruction changes while the robot is moving. Finally, in the \textbf{Vision} stage, we introduce visual observations through gated cross-attention and post-train the language-action backbone on scarce teleoperated data. This design allows visual grounding to build upon an established motion prior rather than learning motion generation from scratch.

Our experiments support this bottom-up approach. The learned action representation preserves motion fidelity and executability, outperforming alternative tokenization methods at comparable bitrates. The language-action model, \method, generates realistic and semantically meaningful motions while achieving substantially stronger execution robustness than recent text-to-motion models~\cite{rempe2026kimodo,zhao2026ardy,huang2026omg}, including under perturbations, extended horizons, and different whole-body trackers. Transition-aware post-training extends the same policy to interactive control, markedly improving how often the robot follows an instruction that changes mid-motion. Finally, vision-stage experiments show that the pretrained language-action backbone improves visually grounded task execution after post-training on teleoperated demonstrations (\cref{tab:humanoidarena,fig:vla}).

\Needspace{4\baselineskip}Our contributions are threefold:
\begin{itemize}
    \setlength\itemsep{0.1em}
    \item We propose a \textbf{bottom-up learning paradigm} for humanoid whole-body intelligence, progressively building from action to language to vision according to data availability and the requirements of physical execution.
    \item We develop an \textbf{execution-grounded motion representation} by curating tracker-verified motions and learning a four-stream residual-VQ tokenizer, providing a compact and reliable action space for whole-body generation.
    \item We demonstrate \textbf{language-conditioned whole-body control} with strong semantic alignment and physical execution across different trackers, and show how transition-aware post-training extends the same policy to robust streaming interaction under changing instructions. Vision-stage experiments further demonstrate that the learned motion prior enhances subsequent VLA training, achieving first-class performance among humanoid models.
\end{itemize}

\section{Related work}
\label{sec:related}

\begin{figure*}[t]
\centering
\vspace{5pt}
\includegraphics[width=\textwidth]{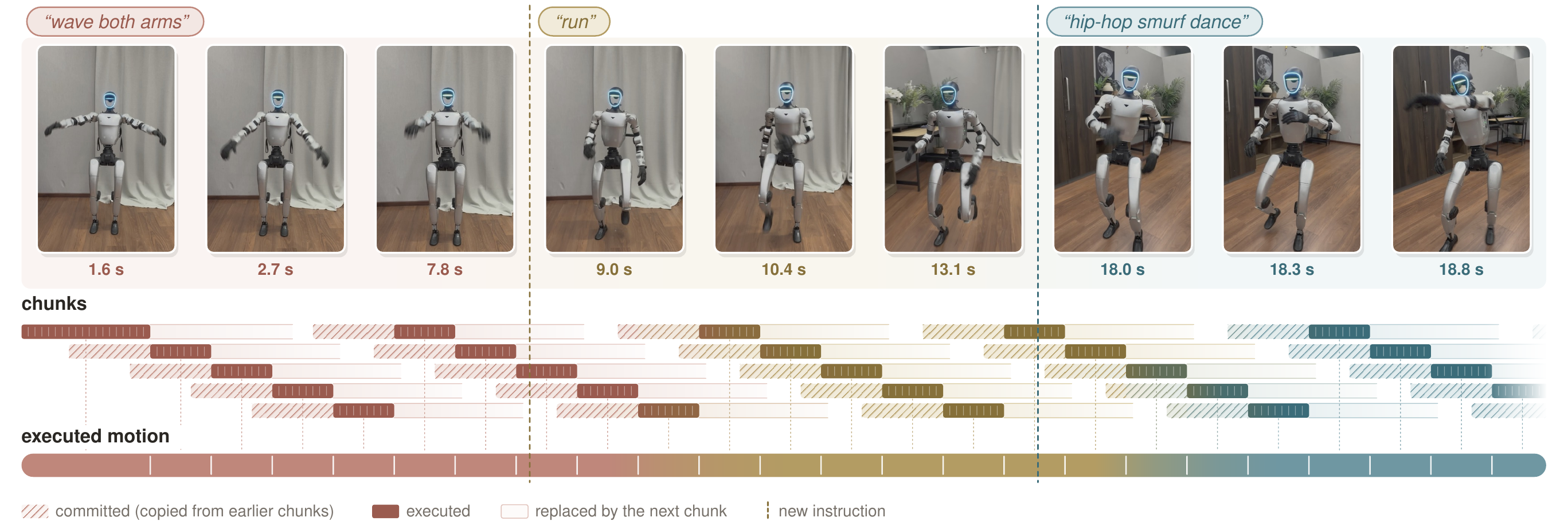}
\caption{\textbf{Interactive whole-body control through rolling motion plans.} Real-robot frames illustrate successive instructions during continuous execution. Each new window preserves committed motion (hatched) while regenerating the future; solid portions are executed and faded predictions are superseded by later plans. Markers indicate instruction changes.}
\label{fig:streaming}
\end{figure*}

\paragraph{Text-driven motion generation.}
Text-to-motion models trained on human motion capture~\cite{guo2022humanml3d,petrovich2022temos,guo2022tm2t,tevet2023mdm,zhang2024motiondiffuse,chen2023mld,zhang2023t2mgpt,guo2024momask,pinyoanuntapong2024mmm,jiang2023motiongpt,dai2024motionlcm,meng2025mardm} established retrieval-based evaluation with learned text--motion encoders (R-precision, FID~\cite{heusel2017fid}, multimodal distance)~\cite{guo2022humanml3d,petrovich2023tmr}. Recent work scales data and models~\cite{lin2023motionx,lu2025scamo,wang2025motionlib,fan2025gotozero,hymotion2025}, generates in a streaming fashion from a causal latent history~\cite{xiao2025motionstreamer}, and injects physics through simulation~\cite{yuan2023physdiff}. For humanoid robots, text-conditioned planners generate reference motion~\cite{jiang2024harmon,rempe2026kimodo,zhao2026ardy,huang2026omg,lighto1,xie2026textop}, commonly separating kinematic generation from execution by a whole-body tracker. In contrast, \method generates in a latent space learned only from tracker-executed motion and scores generated motion after physical execution under three trackers.

\paragraph{Motion tokenization.}
Vector quantization~\cite{vandenoord2017vqvae} with residual~\cite{zeghidour2022soundstream,lee2022rqvae}, product~\cite{jegou2011pq} or finite scalar~\cite{mentzer2024fsq} codebooks underlies discrete motion tokens for language models and masked generators~\cite{guo2022tm2t,zhang2023t2mgpt,jiang2023motiongpt,guo2024momask,lu2024humantomato,zou2024parco,lu2025scamo}. Robot-oriented motion tokens include the FSQ universal token of SONIC~\cite{luo2026sonic} and the multi-head codebooks of MotionBricks~\cite{wang2026motionbricks}. Our tokenizer differs in what it is trained on (tracker rollouts rather than raw capture) and in its four streams for root, legs, arms and foot contact; we compare it with these codecs and with product VQ and FSQ at equal bitrate, both kinematically and after execution.

\paragraph{Humanoid whole-body control.}
Physics-based character controllers support motion tracking and object interaction in simulation~\cite{luo2023phc,xu2025intermimic,tessler2024maskedmimic}. On real humanoids, motion trackers and behavior foundation models execute reference motion~\cite{he2024h2o,he2024omnih2o,fu2024humanplus,cheng2024exbody,ji2026exbody2,he2025asap,he2025hover,chen2025gmt,ze2025twist,xie2025kungfubot,liao2026beyondmimic,luo2026sonic,qi2026humanoidgpt,zeng2026scalebfm,yuan2026bfmsurvey}, usually from human motion retargeted to the robot~\cite{araujo2026gmr,yang2026omniretarget} or produced through iterative data synthesis~\cite{xu2026icms}. Language-conditioned controllers learn from human videos~\cite{mao2025uh1} or tracker rollouts~\cite{shao2025langwbc,wang2026sentinel}, plan with latent vision-language instructions~\cite{xue2025leverb}, or replan motion in closed loop~\cite{chen2026reactivebfm}. \method keeps the planner--tracker interface but makes the planner's output space executable by construction, and it measures closed-loop robustness and dependence on the tracker.

\paragraph{Vision-language-action models.}
Robot policies predict discrete action tokens~\cite{brohan2023rt1,brohan2023rt2,kim2024openvla,pertsch2025fast} or use a diffusion or flow action expert~\cite{chi2023diffusionpolicy,ghosh2024octo,black2025pi0,black2025pi05,bjorck2025gr00t,hou2025dita}, often predicting action chunks~\cite{zhao2023act} executed with real-time chunking~\cite{black2025rtc,black2025trainingrtc}. Humanoid VLAs, WAMs, and vision-based policies extend these ideas to whole-body control~\cite{ding2025humanoidvla,xue2025leverb,jiang2026wholebodyvla,zhang2026wholebodywam,luo2026beingh0,he2026viral,yang2026zerowbc}, and GR00T~N1.7 can emit latent tokens that SONIC decodes~\cite{nvidia2026gr00tn17,luo2026sonic}. Pretraining the action expert before or without vision improves instruction generalization~\cite{lin2026la4vla,xu2026apt}, while a frozen pretrained trajectory backbone can serve many tasks~\cite{zhou2026fss}. Two concurrent works are closest to ours: Kimodo-Policy conditions a frozen text-to-motion model on vision through a ControlNet-style adapter~\cite{zhang2023controlnet,wang2026kimodopolicy}, and Movens-Zero scales execution-verified human-video motion for humanoid foundation models~\cite{bu2026movenszero}. \method instead pretrains the VLA's own trunk on executable whole-body motion with language before vision and post-trains it end to end.

\section{Method}
\label{sec:method}

\subsection{Learning whole-body intelligence in reverse}
\label{sec:formulation}

ALV builds whole-body intelligence from the body up. Rather than learning perception and motor behavior jointly from scarce visual demonstrations, we first establish \emph{how to move}, then learn \emph{what to do} from language, and finally ground that behavior in \emph{where to act} through vision. Each stage builds on the previous one while introducing a richer source of conditioning.

A whole-body tracker $T$~\cite{qi2026humanoidgpt} executes a reference motion $m$, producing the realized trajectory $T(m)$. We define executable references by stable, faithful tracking:
\begin{equation}
\label{eq:executable}
\mathcal{E}=\big\{m:\operatorname{stable}(T(m)),\;
       d_{\mathrm{track}}\big(T(m),m\big)\leq\epsilon\big\},
\end{equation}
where $d_{\mathrm{track}}$ measures discrepancies in articulation, root motion, and contact, and $\epsilon$ is a tracking tolerance. Execution thus shapes the motion data used for learning, rather than serving only as a final evaluation. The three bands of \cref{fig:method} illustrate the progression from an executable \textbf{Action} space, to interactive \textbf{Language} control, to \textbf{Vision}-conditioned behavior.

\subsection{Action: an execution-grounded motion space}
\label{sec:action}

\paragraph{Learning from realized motion.}
We use 2,530 hours of motion data in total, all captured in-house. Although human motion is diverse, retargeted references are not necessarily executable by a humanoid. We therefore retarget candidate motions to the robot~\cite{araujo2026gmr}, execute them with a frozen whole-body tracker in simulation~\cite{todorov2012mujoco}, and retain stable, faithfully tracked segments. Our representation learns from the \emph{realized} trajectories $T(m)$ rather than the original references $m$.

\paragraph{A structured motion vocabulary.}
An encoder $E_\phi$ and residual vector quantizer $Q$~\cite{zeghidour2022soundstream,lee2022rqvae} map accepted trajectories into four streams---root, lower body, upper body, and contact---with stream-specific encoders and codebooks. Latent mixing and a shared temporal decoder $D_\psi$ recover coordinated whole-body motion (band 1 of \cref{fig:method}). Each latent row is
\begin{equation}
\label{eq:motion_codes}
z_t=\big[z_t^{\mathrm{root}};z_t^{\mathrm{lower}};
          z_t^{\mathrm{upper}};z_t^{\mathrm{contact}}\big]
          \in\mathbb{R}^{304}.
\end{equation}
Progressive training combines continuous and quantized reconstruction with kinematic, temporal, and contact objectives. Generation remains continuous: a predicted latent $x$ is projected onto the learned codebooks by $\Pi$ and decoded into motion,
\begin{equation}
\label{eq:motion_decode}
\hat m=D_\psi\big(\Pi(x)\big).
\end{equation}
The resulting vocabulary connects high-level motion generation to low-level execution. For online control, we distill the context-aware decoder into a bounded-lookahead decoder, allowing future motion to be decoded while a committed prefix is being executed.

\subsection{Language: from motion generation to interactive whole-body control}
\label{sec:language}
\label{sec:streaming}

\paragraph{Instruction-conditioned generation.}
With an executable motion space in place, the next challenge is to turn motion into behavior that can be commanded and revised while the robot moves. We use 1,175 hours of data, combining our in-house captures with selected subsets of OMG~\cite{huang2026omg} and MotionMillion~\cite{fan2025gotozero}. A frozen Qwen3-VL-2B-Instruct vision-language backbone~\cite{bai2025qwen3vl} supplies instruction features without image inputs, while a trainable DiT~\cite{peebles2023dit} $f_\theta$ predicts latent motion conditioned on instruction $c$, proprioception $s$, and committed motion~\cite{lipman2023flowmatching}. Given target latents $x_1$, Gaussian noise $n\sim\mathcal{N}(0,I)$, and flow time $\tau$, we form $x_\tau=(1-\tau)n+\tau x_1$ and optimize
\begin{equation}
\label{eq:latent_flow}
\mathcal{L}_{\mathrm{FM}}=\mathbb{E}\!\left[
\left\|M\odot\left(f_\theta(x_\tau,\tau,c,s)
                      -(x_1-n)\right)\right\|_2^2\right],
\end{equation}
where $M$ masks padding and committed rows, which remain fixed as context. Whole-segment training learns the correspondence between instructions and complete behaviors; continuation training teaches the model to extend motion already in progress. Classifier-free guidance~\cite{ho2022cfg} further strengthens instruction conditioning.

\paragraph{Streaming execution.}
At deployment, each predicted chunk is a \emph{revisable plan}, not a trajectory to execute in full. Every $P$ rows, the policy predicts a new $W$-row window from the updated instruction and robot state, retaining $C$ committed rows from the preceding plan while regenerating the future. The robot continues executing that prefix during planning and decoding. With planning delay $d$ and decoder look-ahead $\ell$, measured in rows, uninterrupted replacement requires $d+\ell\leq C$.

\begin{table*}[t]
\centering
\caption{\textbf{Compact motion representation preserves executability.} (a) Matched quantization. (b) Reconstruction and execution; executed MPJPE (mm) measures tracked reconstruction error against the original reference. Ground truth is unranked.}
\label{tab:action}
\footnotesize
\setlength{\tabcolsep}{4.5pt}
\renewcommand{\arraystretch}{1.12}
\phantomsection\label{tab:action-a}\textbf{(a) Quantization at 638 bits per 80\,ms row} (held-out TK-test recordings)\\[3pt]
\fitwidth{%
\begin{tabular}{l cccccccc}
\toprule
 & Joint MAE$\downarrow$ & Keypoint$\downarrow$ & Root drift$\downarrow$ & Root yaw$\downarrow$ & Contact & Jerk ratio & Dead codes$\downarrow$ & Gates$\uparrow$ \\
Quantizer & ($10^{-3}$\,rad) & (mm) & (m) & ($^\circ$) & F1$\uparrow$ & ($\to 1$) & (\%) & (of 47) \\
\midrule
FSQ~\cite{mentzer2024fsq} & 15.79 & 4.52 & 0.101 & 0.307 & 0.9960 & 0.75 & 35 & 39 \\
Residual FSQ & 14.54 & 3.23 & 0.090 & \second{0.177} & 0.9959 & 0.75 & 54 & 40 \\
Product VQ~\cite{jegou2011pq} & \second{3.93} & \second{2.39} & \second{0.078} & 0.182 & \second{0.9989} & \second{0.80} & \best{0} & \second{43} \\
\midrule
\rowcolor{alvOurs}\method\ RVQ (ours) & \best{2.53} & \best{1.44} & \best{0.052} & \best{0.102} & \best{0.9993} & \best{0.81} & \best{0} & \best{46} \\
\bottomrule
\end{tabular}}\\[8pt]
\phantomsection\label{tab:action-b}\textbf{(b) Codec-to-tracker round trip} (encode, decode, and physically execute)\\[3pt]
\fitwidth{%
\begin{tabular}{l c ccc cc cc}
\toprule
 & & \multicolumn{3}{c}{Noitom test} & \multicolumn{2}{c}{LAFAN1-G1} & \multicolumn{2}{c}{OmniRetarget} \\
\cmidrule(lr){3-5}\cmidrule(lr){6-7}\cmidrule(lr){8-9}
 & Latency & Joint & & Exec. & Joint & Exec. & Joint & Exec. \\
Codec & (ms) & MAE$\downarrow$ & R@1$\uparrow$ & MPJPE$\downarrow$ & MAE$\downarrow$ & MPJPE$\downarrow$ & MAE$\downarrow$ & MPJPE$\downarrow$ \\
\midrule
Ground truth & -- & \textcolor{alvGrey}{0.0} & \textcolor{alvGrey}{84.6} & \textcolor{alvGrey}{12.01} & \textcolor{alvGrey}{0.0} & \textcolor{alvGrey}{72.84} & \textcolor{alvGrey}{0.0} & \textcolor{alvGrey}{33.08} \\
\midrule
MotionBricks~\cite{wang2026motionbricks} & 2{,}133 & 104.0 & \second{81.3} & 22.29 & 165.7 & 95.00 & 166.2 & 50.46 \\
SONIC tokens~\cite{luo2026sonic} & 1{,}000 & 93.7 & 65.4 & 43.70 & 157.2 & 143.81 & 145.4 & 110.73 \\
\midrule
\rowcolor{alvOurs}\method\ RVQ (ours) & \second{560} & \best{3.5} & \best{84.7} & \best{12.08} & \best{48.5} & \best{71.55} & \best{23.5} & \best{37.38} \\
\rowcolor{alvOurs}\method\ RVQ-L3 (causal, ours) & \best{240} & \second{7.1} & \best{84.7} & \second{12.78} & \second{63.8} & \second{75.37} & \second{31.0} & \second{40.61} \\
\bottomrule
\end{tabular}}
\end{table*}

\paragraph{Transition-aware post-training.}
Yet temporal continuity alone cannot resolve changing intent: the retained prefix reflects the previous command, while the new suffix must follow the next one. We therefore use transition-aware post-training with instruction-switch chains spanning successive replanning steps, cross-instruction roll-in from the model's own quantized predictions, and short-segment continuations that require redirection before a behavior naturally ends. An instruction change can also trigger earlier replanning while preserving enough committed motion to cover the delay. Together, these mechanisms provide robust streaming interactive control, enabling seamless transitions between committed and newly generated motion while adapting future behavior to changing instructions, without adding model parameters (\cref{fig:streaming}).

\subsection{Vision: grounding motion priors in the scene}
\label{sec:vision}

The final post-training stage brings visual grounding to a policy that has already learned executable, instruction-responsive motion. We condition the pretrained Qwen3-VL-based backbone on head and wrist images together with language, and initialize the action-generating DiT trunk from the Language stage (band 3 of \cref{fig:method}). We adapt the policy to the target robot, including its hands, using additional state and action projections and post-train it to predict native action chunks from observations and proprioception. Thus, scarce visual demonstrations can teach the policy how to apply its motion prior in a scene, instead of requiring it to learn whole-body behavior from scratch.

\section{Experiments}
\label{sec:experiments}

We ask whether a whole-body \emph{action} representation preserves fidelity and executability (\cref{sec:exp-action}) and supports precise, reliable \emph{language} control (\cref{sec:exp-language}). We then test whether the policy can \emph{adapt} to changing instructions (\cref{sec:exp-streaming}) and whether a prior learned without images benefits \emph{vision}-conditioned action (\cref{sec:exp-vision}).

\paragraph{Experimental setup.}
Unless otherwise specified, we execute Unitree G1~\cite{unitreeg1} motions with Humanoid-GPT~\cite{qi2026humanoidgpt} in MuJoCo~\cite{todorov2012mujoco}. Comparisons cover motion codecs, text-to-motion models, and streaming planners. See \hyperref[app:protocol]{Appendix~\ref*{app:protocol}} for benchmarks and metrics.

\subsection{Action: learning a faithful and executable motion space}
\label{sec:exp-action}

\paragraph{A compact space that retains whole-body structure.}
With encoder, decoder, training data, and bitrate matched, \method's quantizer best preserves local pose while retaining root motion and contacts on held-out recordings (\hyperref[tab:action-a]{Tab.~\ref*{tab:action}(a)}). These findings indicate that compact encoding preserves temporal whole-body structure as well as pose fidelity.

\begin{table*}[t]
\centering
\caption{\textbf{Semantic control and physical execution.} \textbf{(a)} Alignment and instruction following. \textbf{(b)} Success and 5$\times$ open-loop fall rate (\%) are the minimum and maximum across Noitom, OMG, and ReactiveBFM under 100\,N random pushes, respectively. Ground truth is unranked.}
\label{tab:main}
\footnotesize
\setlength{\tabcolsep}{3.0pt}
\renewcommand{\arraystretch}{1.07}
\fitwidth{%
\begin{tabular}{l cc cc ccccc ccccc}
\toprule
 & \multicolumn{9}{c}{\textbf{(a) Language--motion alignment and fine-grained instruction following}} & \multicolumn{5}{c}{\textbf{(b) Motion naturalness and physical reliability}} \\
\cmidrule(lr){2-10}\cmidrule(lr){11-15}
 & \multicolumn{2}{c}{Noitom test} & \multicolumn{2}{c}{OMG test} & \multicolumn{5}{c}{Sealed rule-checked suite (\%)} & & & & & \\
\cmidrule(lr){2-3}\cmidrule(lr){4-5}\cmidrule(lr){6-10}
Model & FID$\downarrow$ & R@3$\uparrow$ & FID$\downarrow$ & R@3$\uparrow$ & Prim.$\uparrow$ & Gest.$\uparrow$ & Lat.$\uparrow$ & Order$\uparrow$ & Loco.$\uparrow$ & Jerk ($\to$GT) & 1$\times$ open$\uparrow$ & 5$\times$ open$\uparrow$ & 5$\times$ closed$\uparrow$ & Max fall$\downarrow$ \\
\midrule
Ground truth & -- & \textcolor{alvGrey}{97.4} & -- & \textcolor{alvGrey}{90.2} & -- & -- & -- & -- & -- & \textcolor{alvGrey}{42.5} & \textcolor{alvGrey}{92.2} & -- & -- & -- \\
\midrule
OMG-500M~\cite{huang2026omg} & 0.589 & 41.7 & 0.535 & \second{65.1} & 26.2 & 43.8 & 27.0 & 7.8 & 37.5 & \best{38.6} & \second{95.6} & 63.0 & 38.8 & 9.2 \\
Kimodo-G1~\cite{rempe2026kimodo} & 0.311 & \second{82.2} & 0.482 & 45.5 & \second{57.7} & \second{84.4} & \second{55.1} & \second{33.6} & \best{85.9} & 120.6 & 89.8 & 8.2 & -- & 63.2 \\
ARDY-G1~\cite{zhao2026ardy} & 0.284 & 70.3 & \second{0.467} & 47.8 & 46.5 & 78.5 & 53.1 & 7.8 & \second{79.3} & 91.7 & 91.3 & \second{86.6} & -- & \second{2.2} \\
TextOp~\cite{xie2026textop} & 0.525 & 31.2 & 0.552 & 24.2 & 11.6 & 24.2 & 6.6 & 3.9 & 51.2 & 393.3 & 92.6 & 79.6 & -- & 5.8 \\
\midrule
\rowcolor{alvOurs}ALV\ (ours) & \best{0.088} & \best{90.4} & \best{0.157} & \best{85.0} & \best{63.4} & \best{84.8} & \best{68.0} & \best{37.5} & 75.0 & \second{46.0} & \best{98.9} & \best{98.8} & \best{98.6} & \best{0.0} \\
\bottomrule
\end{tabular}}
\end{table*}

\begin{table*}[!t]
\centering
\caption{\textbf{Execution under three trackers.} Success (\%) at 5$\times$ durations with 100\,N pushes; the lower block replans in closed loop. \method\ runs in the deployment setting. At 1$\times$, \method\ succeeds in 99.1--100\% of episodes in all nine columns. TextOp is evaluated with Humanoid-GPT only.}
\label{tab:trackers}
\footnotesize
\setlength{\tabcolsep}{6pt}
\begin{tabular}{l ccc ccc ccc}
\toprule
 & \multicolumn{3}{c}{Humanoid-GPT~\cite{qi2026humanoidgpt}} & \multicolumn{3}{c}{SONIC~\cite{luo2026sonic}} & \multicolumn{3}{c}{ScaleBFM~\cite{zeng2026scalebfm}} \\
\cmidrule(lr){2-4}\cmidrule(lr){5-7}\cmidrule(lr){8-10}
Model & Noitom & OMG & RBFM & Noitom & OMG & RBFM & Noitom & OMG & RBFM \\
\midrule
OMG-500M~\cite{huang2026omg} & 78.4 & 80.9 & 63.0 & 82.1 & 90.1 & 79.4 & 86.2 & 87.6 & 81.8 \\
Kimodo-G1~\cite{rempe2026kimodo} & 49.0 & \second{91.8} & 8.2 & 50.1 & \second{93.9} & 10.1 & 49.6 & \second{95.0} & 8.4 \\
ARDY-G1~\cite{zhao2026ardy} & 90.4 & 87.1 & \second{86.6} & \second{97.1} & 93.3 & \second{94.6} & \second{97.1} & 93.4 & \second{93.9} \\
TextOp~\cite{xie2026textop} & \second{96.9} & 79.6 & 79.8 & -- & -- & -- & -- & -- & -- \\
\rowcolor{alvOurs}\method & \best{100.0} & \best{99.8} & \best{98.7} & \best{99.8} & \best{99.4} & \best{99.0} & \best{99.7} & \best{99.5} & \best{99.4} \\
\midrule
OMG-500M, closed loop & 81.5 & 38.8 & 48.7 & 82.7 & 43.9 & 51.7 & 60.2 & 37.4 & 27.4 \\
\rowcolor{alvOurs}\method, closed loop & \best{100.0} & \best{99.8} & \best{98.4} & \best{99.6} & \best{99.7} & \best{98.9} & \best{99.6} & \best{99.6} & \best{99.4} \\
\bottomrule
\end{tabular}
\end{table*}

\paragraph{The decisive test is execution, not reconstruction.}
On Noitom validation, decoded \method\ motion closely matches direct reference tracking while preserving language alignment (\hyperref[tab:action-b]{Tab.~\ref*{tab:action}(b)}). MotionBricks and SONIC execute less faithfully despite receiving ground-truth root trajectories; ours receives only an initial anchor.

On LAFAN1-G1~\cite{harvey2020lafan1,unitreelafan} and OmniRetarget~\cite{yang2026omniretarget}, execution remains close to reference tracking, extending the evidence to unseen dynamic and object-interaction motions. The causal decoder also maintains faithful execution on familiar motion with less look-ahead, supporting online control.

\paragraph{What does the learned space contribute to generation?}
Under matched data and training/selection protocols, raw-motion generation follows sealed instructions comparably to \method\ but produces less smooth motion and more foot sliding. The learned representation therefore improves motion quality without sacrificing semantic expressiveness.

\subsection{Language-conditioned control: translating instructions into reliable whole-body motion}
\label{sec:exp-language}
\label{sec:exp-trackers}

\paragraph{Language alignment survives the executable bottleneck.}
On Noitom, \method\ aligns motion with language better than the compared baselines (\cref{tab:main}). The advantage also holds under OMG's independent evaluator, showing that an executable motion representation remains compatible with language alignment across evaluator representations.

\paragraph{Language control extends to motion details.}
On unseen sealed requests, \method\ leads the primary instruction-following comparison, with a particularly clear advantage in left--right control. Its aggregate gains over ARDY, OMG, and TextOp are statistically significant, supporting control over concrete language constraints beyond merely generating plausible actions.

\paragraph{The strongest gains appear after the motion is executed.}
Under the ReactiveBFM protocol~\cite{chen2026reactivebfm}, \method\ produces near-reference smoothness and remains reliable under pushes from nominal through prolonged execution, both in open loop and with closed-loop replanning. It substantially outperforms the strongest prolonged open-loop baseline. Its reported prolonged open-loop tests remain fall-free.

\paragraph{Executability transfers across whole-body trackers.}
Reliability persists across the evaluated Humanoid-GPT, SONIC~\cite{luo2026sonic}, and ScaleBFM~\cite{zeng2026scalebfm} trackers and test sets, including prolonged closed-loop execution under pushes (\cref{tab:trackers}). Language semantics largely survive tracking on the development suite, and executed motion remains competitive under OMG's evaluator. This suggests that the learned prior is usable beyond a single tracking system.

\subsection{Interactive language control: adapting whole-body motion to changing instructions}
\label{sec:exp-streaming}

\begin{table}[t]
\centering
\caption{\textbf{Following new instructions during motion.} Closed-loop scores use tracked motion 0.56--4\,s after each switch. Interpret residue alongside new-command success. $^*$: improvement over \alvl\ under the same schedule; paired 95\% bootstrap CI excludes zero. Reference segments form the unranked GT ceiling.}
\label{tab:streambench}
\footnotesize
\setlength{\tabcolsep}{3.4pt}
\renewcommand{\arraystretch}{1.09}
\fitwidth{%
\begin{tabular}{l cc ccc}
\toprule
 & \multicolumn{2}{c}{Rule success$\uparrow$} & \multicolumn{3}{c}{Periodic requests} \\
\cmidrule(lr){2-3}\cmidrule(lr){4-6}
Model & Periodic & Interrupt & R@1$\uparrow$ & R@3$\uparrow$ & Residue$\downarrow$ \\
\midrule
GT ceiling & \textcolor{alvGrey}{88.6} & \textcolor{alvGrey}{88.6} & \textcolor{alvGrey}{96.9} & \textcolor{alvGrey}{98.6} & -- \\
\midrule
TextOp~\cite{xie2026textop} & 23.9 & -- & 21.3 & 33.0 & \best{15.4} \\
ARDY-G1~\cite{zhao2026ardy} & 25.1 & -- & 25.2 & 33.8 & \second{25.0} \\
\midrule
\rowcolor{alvOurs}\alvl\ (ours) & 26.6 & 35.8 & 27.2 & 40.4 & 34.4 \\
\rowcolor{alvOurs}\alvls\ (ours) & \best{40.0}$^*$ & \best{48.8}$^*$ & \best{37.3}$^*$ & \best{53.6}$^*$ & 26.6$^*$ \\
\bottomrule
\end{tabular}}
\end{table}

\begin{table*}[!t]
\centering
\caption{\textbf{Whole-body visuomotor control on HumanoidArena.} Success rate (\%) with one policy per task, executed by SONIC; AFR: average fall rate; Overall: mean over the seven tasks. Leaderboard entries are copied from the official HumanoidArena leaderboard~\cite{wang2026humanoidarena}; ``--'': not reported. Bottom block: identical VLM backbone, data, and recipe, varying only the action-model initialization; final checkpoints, no selection. $\Delta$: ours minus random initialization (points). Best and second-best per column are bold and underlined.}
\label{tab:humanoidarena}
\footnotesize
\setlength{\tabcolsep}{4.0pt}
\renewcommand{\arraystretch}{1.08}
\fitwidth{%
\begin{tabular}{l c ccc c cccc c c}
\toprule
 & & \multicolumn{4}{c}{Human--object interaction (HOI)} & \multicolumn{5}{c}{Human--scene interaction (HSI)} & \\
\cmidrule(lr){3-6}\cmidrule(lr){7-11}
Method & AFR$\downarrow$ & Football & DoubleDesk & P\&PBox & Avg$\uparrow$ & OpenDoor & SitSofa & Boxing & VisNavi & Avg$\uparrow$ & Overall$\uparrow$ \\
\midrule
\multicolumn{12}{l}{\textit{HumanoidArena leaderboard~\cite{wang2026humanoidarena} (SONIC entries):}} \\
ACT~\cite{zhao2023act} & 8.6 & 16.7 & 18.3 & 56.7 & 30.6 & 78.3 & 73.3 & 56.7 & 33.3 & 60.4 & 47.6 \\
DP~\cite{chi2023diffusionpolicy} & 8.3 & 45.0 & 36.7 & 75.0 & 52.2 & 85.0 & 78.3 & 76.7 & 23.3 & 65.8 & 60.0 \\
FM~\cite{lipman2023flowmatching} & 5.7 & 13.3 & 38.3 & 73.3 & 41.7 & 70.0 & 15.0 & 70.0 & 38.3 & 48.3 & 45.5 \\
$\pi_{0.5}$~\cite{black2025pi05} & \second{5.2} & 10.0 & 43.3 & 71.7 & 41.7 & 66.7 & 73.3 & 70.0 & 23.3 & 58.3 & 51.2 \\
HuGe-OFT~\cite{liu2026huge} & -- & 48.3 & 0.0 & 53.3 & 33.9 & 75.0 & 55.0 & 73.3 & 73.3 & 69.2 & 54.1 \\
HuGe-$\pi$~\cite{liu2026huge} & -- & 63.3 & 11.7 & 76.7 & 50.6 & 88.3 & 83.3 & 68.3 & 63.3 & 75.8 & 65.0 \\
HuGe-GR00T~\cite{liu2026huge} & -- & \best{71.7} & 15.0 & 81.7 & 56.1 & \second{95.0} & 78.3 & 68.3 & 56.7 & 74.6 & 66.7 \\
WB-WAM~\cite{qin2026wbwam} & -- & \second{70.0} & \second{65.0} & \best{86.7} & \best{73.9} & \best{98.3} & \best{95.0} & \second{81.7} & \best{76.7} & \best{87.9} & \best{81.9} \\
\midrule
\multicolumn{12}{l}{\textit{Same VLM backbone, action model initialized from:}} \\
Random init. & 5.5 & 33.3 & 46.7 & 45.0 & 41.7 & 88.3 & 83.3 & 71.7 & 68.3 & 77.9 & 62.4 \\
\rowcolor{alvOurs}\method\ (ours) & \best{3.8} & 60.0 & \best{68.3} & \second{83.3} & \second{70.5} & 90.0 & \second{93.3} & \best{83.3} & \second{75.0} & \second{85.4} & \second{79.0} \\
$\Delta$ (ours $-$ random init.) & \textcolor{alvV}{$-$1.7} & \textcolor{alvV}{+26.7} & \textcolor{alvV}{+21.7} & \textcolor{alvV}{+38.3} & \textcolor{alvV}{+28.9} & \textcolor{alvV}{+1.7} & \textcolor{alvV}{+10.0} & \textcolor{alvV}{+11.7} & \textcolor{alvV}{+6.7} & \textcolor{alvV}{+7.5} & \textcolor{alvV}{+16.7} \\
\bottomrule
\end{tabular}}
\end{table*}

\begin{table*}[!t]
\centering
\caption{\textbf{Whole-body visuomotor control on SIMPLE.} (a) Success rate (\%) over L0--L2, ten episodes per task and level. Means require all tasks in the group. A dash marks an unreported score. For methods marked $^\dagger$, the four additional tasks use \href{https://arxiv.org/html/2610.31166v2}{Chen et al.}'s evaluation. (b) Success counts over all four levels, including Hard. Bold and underline rank the displayed method scores in each column, including ties. The shaded row gives the final episode outcomes for \method. $\Delta$ compares the reported outcomes for \method and random initialization, in percentage points in (a) and successful episodes in (b).}
\label{tab:simple}
\footnotesize
\setlength{\tabcolsep}{3.5pt}
\renewcommand{\arraystretch}{.90}
\textbf{(a) Success rates at L0--L2}\\[2pt]
\fitwidth{%
\begin{tabular}{@{}l ccccc r cccc r@{}}
\toprule
 & \multicolumn{6}{c}{Five core tasks} & \multicolumn{5}{c}{Four additional tasks} \\
\cmidrule(lr){2-7}\cmidrule(l){8-12}
Method & \shortstack{X-move\\pick} & \shortstack{Bend\\pick} & Handover & \shortstack{Pick between\\tables} & \shortstack{Tabletop\\grasp} & Mean$\uparrow$ & \shortstack{Close\\door} & \shortstack{Open\\oven} & \shortstack{Open\\faucet} & \shortstack{Hug\\container} & Mean$\uparrow$ \\
\midrule
\multicolumn{12}{l}{\textit{\href{https://github.com/physical-superintelligence-lab/SIMPLE}{SIMPLE release}}} \\
$\Psi_0$ & 86.7 & \best{100.0} & 80.0 & 60.0 & 93.3 & 84.0 & \best{100.0} & 53.3 & 33.3 & 53.3 & 60.0 \\
DreamZero & \best{100.0} & 86.7 & 80.0 & 36.7 & 86.7 & 78.0 & -- & -- & -- & -- & -- \\
ACT$^{\dagger}$ & 83.3 & 93.3 & 46.7 & 60.0 & 93.3 & 75.3 & \best{100.0} & 6.7 & 3.3 & 50.0 & 40.0 \\
OpenPI $\pi_{0.5}^{\dagger}$ & 43.3 & 93.3 & 46.7 & 30.0 & 93.3 & 61.3 & \second{96.7} & 43.3 & 20.0 & 20.0 & 45.0 \\
GR00T N1.6 & 90.0 & 66.7 & 23.3 & 0.0 & 83.3 & 52.7 & -- & -- & -- & -- & -- \\
Diffusion Policy$^{\dagger}$ & 26.7 & 80.0 & 30.0 & 13.3 & 83.3 & 46.7 & \best{100.0} & 0.0 & 36.7 & 0.0 & 34.2 \\
EgoVLA$^{\dagger}$ & 10.0 & 66.7 & 23.3 & 0.0 & 90.0 & 38.0 & 93.3 & 50.0 & 0.0 & 3.3 & 36.7 \\
InternVLA-M1$^{\dagger}$ & 0.0 & 33.3 & 0.0 & 0.0 & 0.0 & 6.7 & 0.0 & 73.3 & 0.0 & 0.0 & 18.3 \\
H-RDT & 6.7 & 3.3 & 3.3 & 0.0 & 0.0 & 2.7 & -- & -- & -- & -- & -- \\
\midrule
\multicolumn{12}{l}{\textit{Additional per-task policies}} \\
Holo-M \href{https://horizonrobotics.github.io/gail/Holo-M/}{[Shao et al.]} & \best{100.0} & 93.3 & \second{93.3} & \second{83.3} & 90.0 & \second{92.0} & -- & -- & -- & -- & -- \\
WholeBodyMem \href{https://arxiv.org/html/2610.31166v2}{[Chen et al.]} & 90.0 & \best{100.0} & \second{93.3} & 70.0 & \second{96.7} & 90.0 & \best{100.0} & 90.0 & \best{66.7} & 60.0 & \second{79.2} \\
Kimodo-Policy$^{*}$ \href{https://arxiv.org/html/2611.11554v1}{[Wang et al.]} & -- & -- & 83.3 & 70.0 & -- & -- & \best{100.0} & 56.7 & 53.3 & -- & -- \\
$\lambda_0$ \href{https://arxiv.org/html/2610.18950v1}{[Lambda-humanoid Team]} & 93.3 & \best{100.0} & \best{96.7} & 76.7 & 90.0 & 91.3 & -- & -- & -- & -- & -- \\
\midrule
\multicolumn{12}{l}{\textit{Per-task policies evaluated by \href{https://arxiv.org/html/2611.04839v1}{Dong et al.}}} \\
DiT4DiT & 93.3 & 93.3 & \second{93.3} & 70.0 & 86.7 & 87.3 & -- & \best{96.7} & \second{60.0} & 76.7 & -- \\
Fast-WAM & 86.7 & 13.3 & 60.0 & 0.0 & 80.0 & 48.0 & -- & 6.7 & 10.0 & 0.0 & -- \\
VLA-JEPA & 0.0 & 3.3 & 80.0 & 76.7 & 73.3 & 46.7 & -- & 80.0 & 40.0 & 76.7 & -- \\
Action-conditioned latent foresight & 90.0 & \second{96.7} & 90.0 & \best{96.7} & \best{100.0} & \best{94.7} & -- & \best{96.7} & 56.7 & \second{86.7} & -- \\
\midrule
\multicolumn{12}{l}{\textit{Multi-task policies}} \\
Holo-M \href{https://horizonrobotics.github.io/gail/Holo-M/}{[Shao et al.]} & 63.3 & 86.7 & 80.0 & 76.7 & 83.3 & 78.0 & -- & -- & -- & -- & -- \\
Being-M0.7 \href{https://arxiv.org/html/2610.11283v1}{[Yue et al.]} & 93.3 & 73.3 & \second{93.3} & 43.3 & 93.3 & 79.3 & -- & -- & -- & -- & -- \\
StarVLA QwenPI \href{https://huggingface.co/StarVLA/Qwen3-VL-4B-PI-SIMPLE}{[model card]} & \second{96.7} & 66.7 & \second{93.3} & 63.3 & 86.7 & 81.3 & \best{100.0} & 90.0 & \best{66.7} & 66.7 & \best{80.8} \\
\midrule
\multicolumn{12}{l}{\textit{Released $\Psi_0$ checkpoints}} \\
$\Psi_0$, per-task & \best{100.0} & \second{96.7} & 90.0 & 76.7 & 93.3 & 91.3 & 93.3 & 36.7 & 56.7 & 40.0 & 56.7 \\
$\Psi_0$, multi-task & 10.0 & 6.7 & 80.0 & 50.0 & 40.0 & 37.3 & 93.3 & 90.0 & 40.0 & 43.3 & 66.7 \\
\midrule
\multicolumn{12}{l}{\textit{Random-initialization reference}} \\
Random init. & 43.3 & 16.7 & 60.0 & 30.0 & 50.0 & 40.0 & 90.0 & 40.0 & 30.0 & 60.0 & 55.0 \\
\midrule
\multicolumn{12}{l}{\textit{Final episode outcomes}} \\
\rowcolor{alvOurs}\method\ (ours) & \best{100.0} & 26.7 & 56.7 & 46.7 & 86.7 & 63.3 & \best{100.0} & \second{93.3} & 40.0 & \best{90.0} & \best{80.8} \\
$\Delta$ (ours $-$ random init.) & \textcolor{alvV}{+56.7} & \textcolor{alvV}{+10.0} & $-$3.3 & \textcolor{alvV}{+16.7} & \textcolor{alvV}{+36.7} & \textcolor{alvV}{+23.3} & \textcolor{alvV}{+10.0} & \textcolor{alvV}{+53.3} & \textcolor{alvV}{+10.0} & \textcolor{alvV}{+30.0} & \textcolor{alvV}{+25.8} \\
\bottomrule
\end{tabular}}\par\medskip
\textbf{(b) Successes across L0, L1, L2 and Hard}\\[2pt]
\fitwidth{%
\begin{tabular}{@{}l ccccc cccc r@{}}
\toprule
Method & \shortstack{X-move\\pick} & \shortstack{Bend\\pick} & Handover & \shortstack{Pick between\\tables} & \shortstack{Tabletop\\grasp} & \shortstack{Close\\door} & \shortstack{Open\\oven} & \shortstack{Open\\faucet} & \shortstack{Hug\\container} & Total \\
\midrule
$\Psi_0$, per-task & \second{38} & \best{32} & \best{36} & \best{27} & \best{29} & \second{38} & 11 & \best{21} & 13 & \best{245/360} \\
$\Psi_0$, multi-task & 4 & 2 & \second{32} & 17 & 12 & \second{38} & \best{33} & 16 & 13 & 167/360 \\
Random init. & 21 & 5 & 23 & 18 & \second{15} & 30 & \second{21} & 12 & \second{20} & 165/360 \\
\midrule
\rowcolor{alvOurs}\method\ (ours) & \best{39} & \second{8} & 24 & \second{19} & \best{29} & \best{40} & \best{33} & \second{18} & \best{29} & \second{239/360} \\
$\Delta$ (ours $-$ random init.) & \textcolor{alvV}{+18} & \textcolor{alvV}{+3} & \textcolor{alvV}{+1} & \textcolor{alvV}{+1} & \textcolor{alvV}{+14} & \textcolor{alvV}{+10} & \textcolor{alvV}{+12} & \textcolor{alvV}{+6} & \textcolor{alvV}{+9} & \textcolor{alvV}{+74} \\
\bottomrule
\end{tabular}}

\end{table*}

\begin{figure*}[!t]
\centering
\includegraphics[width=\textwidth]{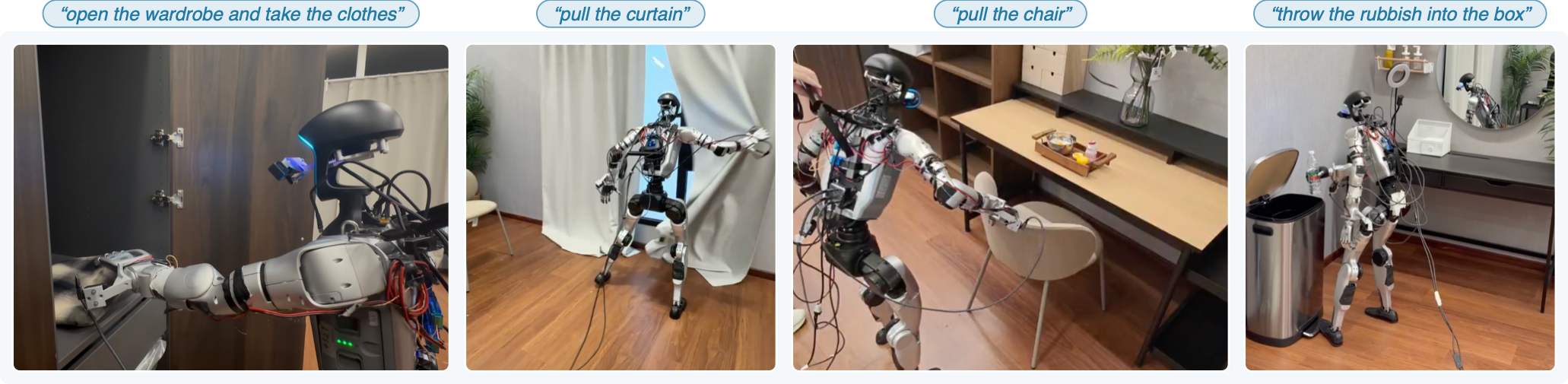}
\caption{Grounding instructions in real scenes. After vision post-training on teleoperated demonstrations, ALV controls a physical G1 from head and wrist images, proprioception, and the instruction shown above each frame. Frames are snapshots from real-robot tests.}
\label{fig:vla}
\end{figure*}

\paragraph{Changing intent is different from generating another clip.}
Here, \alvl\ denotes the single-instruction ALV model and \alvls\ its streaming variant. Interactive control must reconcile changing instructions with motion already committed to execution. In closed-loop tests, we assess whether \emph{executed} motions adopt the new intent while reducing persistence of the previous command (\cref{tab:streambench}). \hyperref[app:exp-streaming]{Appendix~\ref*{app:exp-streaming}} compares state inputs, update schedules and response latencies across planners.

\paragraph{Streaming post-training teaches motion transitions.}
Under periodic requests, \alvls\ follows changed instructions significantly better than the single-instruction \alvl\ without adding parameters and also outperforms ARDY and TextOp. It improves alignment and reduces old-command residue relative to \alvl. Gains are particularly clear when the new intent opposes ongoing motion, suggesting better transitions between behaviors.

On held-out recordings, \alvls\ significantly improves transitions to changed motion directions at natural caption boundaries without new text prompts. Nominal closed-loop execution remains fall-free and better aligned than \alvl, supporting improved connections between motions beyond responsiveness to explicit instruction changes.

\paragraph{Better transitions and earlier requests are complementary.}
Interrupt scheduling significantly improves \alvls\ instruction following over periodic requests and enables earlier replanning; switch training significantly improves instruction satisfaction under either schedule. Responsiveness therefore depends on both \emph{when} replanning occurs and \emph{how} the model connects ongoing motion to the new intent.

\subsection{Vision: transferring an action prior learned without images}
\label{sec:exp-vision}

\paragraph{HumanoidArena.}
On HumanoidArena~\cite{wang2026humanoidarena}, ALV's action initialization yields an overall gain over random initialization under matched VLM, data, and training conditions, and also achieves first-class performance among community baselines (\cref{tab:humanoidarena}). Policies are trained per task and executed with SONIC~\cite{luo2026sonic}; the gains extend across all evaluated tasks. These results suggest that an action prior learned without images can support visually grounded control.

\paragraph{SIMPLE.}
We evaluate nine tasks on the simulated G1 using the public demonstrations from \href{https://arxiv.org/abs/2606.08278}{SIMPLE}. Seven tasks use teleoperated demonstrations and two use motion-planned demonstrations. Each difficulty level contains ten episodes per task. L0 varies surface appearance and distractors, L1 adds lighting changes and L2 also varies the initial target pose. Across these levels, \method completes 95/150 episodes on the five core tasks and 97/120 on the four additional tasks. Across all four levels, including Hard, \method completes 239/360 episodes (66.4\%) and random initialization completes 165/360 (45.8\%), a difference of 74 successes (\cref{tab:simple}).

The additional baselines include per-task and multi-task policies. The latent foresight policy receives demonstration subgoals, $\lambda_0$ uses task-specific configurations, and StarVLA uses its instruction-aligned X-move pick evaluation.

\paragraph{Physical G1.}
We have also done vision post-training on 11.1 hours of in-house teleoperated demonstrations across 4 tasks, ALV controls a physical G1 using head and wrist images, proprioception, and instructions. The snapshots in \cref{fig:vla} illustrate this instruction-conditioned behavior during real-robot tests in household scenes.



\section{Conclusion}
\label{sec:conclusion}

We presented \method{}, a bottom-up approach to humanoid whole-body intelligence that progressively learns how to move, what to do, and where to act. Unlike conventional vision-first approaches, \method{} establishes reliable motor capabilities before introducing language and visual conditioning, allowing each stage to learn from the data most readily available to it. Central to this approach is an execution-grounded motion representation learned from trajectories that a whole-body tracker can faithfully execute, rather than from human motion alone. Built upon this foundation, our language-conditioned policy generates realistic and semantically meaningful whole-body behaviors while maintaining strong execution robustness across different trackers, external perturbations, and extended motion horizons. Transition-aware post-training further enables continuous control under changing instructions, allowing the robot to revise its future motion without disrupting ongoing execution. Finally, vision post-training transfers these learned capabilities to visually grounded tasks, improving performance over random action initialization and demonstrating instruction-conditioned behavior on a physical humanoid. Although complex, multi-step instruction following remains an open challenge, our findings highlight a broader principle: for humanoids, physical executability is not merely a constraint on generated actions, but a foundation upon which higher-level intelligence can be built. By learning motion first and progressively grounding it in language and vision, \method{} offers a promising path toward scalable and reliable whole-body intelligence.

\label{main:end}
\FloatBarrier
\bibliographystyle{assets/plainnat}
\bibliography{main}

\clearpage
\appendix
\twocolumn[{\centering\LARGE\sffamily\bfseries Supplementary Material\par\medskip}]
\setcounter{table}{0}
\setcounter{figure}{0}
\renewcommand{\thetable}{S\arabic{table}}
\renewcommand{\thefigure}{S\arabic{figure}}
\providecommand{\theHtable}{}\providecommand{\theHfigure}{}
\renewcommand{\theHtable}{supp.\arabic{table}}
\renewcommand{\theHfigure}{supp.\arabic{figure}}

\crefname{appendix}{Sec.}{Secs.}
\Crefname{appendix}{Section}{Sections}
\crefname{subappendix}{Sec.}{Secs.}
\Crefname{subappendix}{Section}{Sections}
\newcolumntype{Y}{>{\raggedright\arraybackslash}X}
\newcolumntype{Z}[1]{>{\raggedright\arraybackslash}p{#1}}
\providecommand{\suppline}[2]{\noindent\hyperref[#1]{#2}\dotfill\pageref{#1}\par}

\vspace{4pt}\noindent\textbf{Contents}\par\smallskip
{\small
\suppline{app:data}{\textbf{A\quad Data}}
\suppline{app:curation}{\quad A.1\enspace Motion data and execution-grounded curation}
\suppline{app:stagedata}{\quad A.2\enspace Training and evaluation splits}
\suppline{app:model}{\textbf{B\quad Model}}
\suppline{app:tokenizer}{\quad B.1\enspace Action: motion tokenizer}
\suppline{app:language}{\quad B.2\enspace Language: flow-matching policy}
\suppline{app:switch}{\quad B.3\enspace Language: transition-aware post-training}
\suppline{app:vision-model}{\quad B.4\enspace Vision: post-training on demonstrations}
\suppline{app:exp}{\textbf{C\quad Experiments}}
\suppline{app:protocol}{\quad C.1\enspace Benchmarks and metrics}
\suppline{app:exp-action}{\quad C.2\enspace Effect of decoder look-ahead}
\suppline{app:exp-raw}{\quad C.3\enspace Raw and learned action spaces}
\suppline{app:exp-push}{\quad C.4\enspace Execution under strong pushes}
\suppline{app:exp-streaming}{\quad C.5\enspace SwitchBench and natural transitions}
\suppline{app:exp-real}{\quad C.6\enspace Vision: real-robot post-training}
}

\noindent In the result tables, bold and underline mark the best and second-best distinct values within each comparison, respectively. Tied scores share a rank.

\section{Data}
\label{app:data}

\begin{table*}[t]
\centering
\caption{\textbf{Training data and evaluation splits by stage.} Motion is split by recording before windowing.}
\label{tab:stagedata}
\footnotesize
\renewcommand{\arraystretch}{1.12}
\begin{tabularx}{\textwidth}{@{}l l Y Z{3.6cm}@{}}
\toprule
Stage & Component or benchmark & Training data & Evaluation split \\
\midrule
Action & Motion tokenizer & 2{,}530\,h of in-house motion capture. The tracker executes retargeted motions for tokenizer training. & 791 validation and 1{,}696 test recordings \\
Language & Motion policy & 1{,}175\,h of captioned motion from in-house capture and subsets of OMG~\cite{omgdata} and MotionMillion~\cite{fan2025gotozero}. & 174 Noitom validation and 156 test recordings \\
 & Transition training & Caption boundaries from the language corpus. & 476 held-out recordings for natural transitions \\
\midrule
Vision & Physical G1 & 11.1\,h of teleoperation across four household tasks, with head and wrist cameras. & Physical trials on four household tasks \\
 & HumanoidArena & 100 SONIC demonstrations per task across seven tasks~\cite{wang2026humanoidarena}. & Official evaluation scenes \\
 & SIMPLE & Public demonstrations from seven teleoperated and two motion-planned tasks. & 10 episodes per task at each of L0, L1, L2 and Hard \\
\bottomrule
\end{tabularx}
\end{table*}

\subsection{Motion data and execution-grounded curation}
\label{app:curation}

\paragraph{Tracking and acceptance.}
We retarget in-house recordings to the G1 with GMR~\cite{araujo2026gmr}, resample at 50\,Hz and execute them in MuJoCo~\cite{todorov2012mujoco} with Humanoid-GPT~\cite{qi2026humanoidgpt}. The tokenizer trains only on executed trajectories that satisfy Eq.~(\ref{eq:executable}), \ie, that closely track the reference in joints, keypoints and root motion and keep stable foot contact.
\paragraph{Captions.}
Captions come from the sources where available. Our in-house recordings are segmented with SMQ~\cite{gokay2025smq}, trained on BABEL~\cite{punnakkal2021babel}, and captioned with structured motion descriptions~\cite{zhang2026smd} by a LoRA-tuned~\cite{hu2022lora} Qwen2.5-7B~\cite{qwen2025qwen25}. Clips whose captions mention props or support objects are kept as motion only.

\subsection{Training and evaluation splits}
\label{app:stagedata}
\Cref{tab:stagedata} lists the training sources and held-out splits. External reconstruction tests use LAFAN1-G1~\cite{harvey2020lafan1,unitreelafan} and OmniRetarget~\cite{yang2026omniretarget}. Text-to-motion evaluation uses Noitom, OMG's released test split~\cite{huang2026omg} and the 100STYLE pairs used by ReactiveBFM~\cite{mason2022style100,chen2026reactivebfm}. The development and sealed instruction suites are disjoint. In SIMPLE, we evaluate the simulated G1 over 270 episodes at L0--L2 and 360 including Hard (\cref{tab:simple}). \Cref{tab:benchmarks} lists the metrics.

\section{Model}
\label{app:model}

\subsection{Action: motion tokenizer}
\label{app:tokenizer}

\paragraph{Structure.}
The root, lower body, upper body and foot-contact streams have separate residual codebooks. Their latent dimensions are 64, 96, 128 and 16, respectively. Together they encode each 80\,ms row in 65 codes (638 bits). A shared temporal decoder reconstructs whole-body motion.

\paragraph{Training and decoding.}
Tokenizer training combines reconstruction, kinematic, temporal and contact losses. The 47 test checks in \cref{tab:action} cover reconstruction accuracy, motion dynamics, nearly static motion, codebook usage and export consistency. We distill a bounded-look-ahead decoder from the seven-row context-aware decoder while keeping the encoder and codebooks fixed. \Cref{tab:lookahead} compares the two decoders on identical codes.

\paragraph{Codebook ablation.}
At comparable bitrate, separate streams reduce joint MAE from 3.18 to 2.53 compared with a shared codebook. Removing the contact stream reduces contact F1 from 0.9993 to 0.9962. Smaller codebooks require more predictions per row (\cref{tab:ablation}).

\begin{table}[t]
\centering
\caption{\textbf{Tokenizer design choices on TK-test.} Variants use comparable bitrates. Codes/row gives the predictions required per 80\,ms. Joint MAE in $10^{-3}$\,rad.}
\label{tab:ablation}
\footnotesize
\setlength{\tabcolsep}{3.0pt}
\fitwidth{%
\begin{tabular}{l cc cccc}
\toprule
 & & Codes & Joint & Keypoint & Drift & Contact \\
Variant & Bits & /row$\downarrow$ & MAE$\downarrow$ & (mm)$\downarrow$ & (m)$\downarrow$ & F1$\uparrow$ \\
\midrule
\rowcolor{alvOurs}\method (4 streams) & 638 & \second{65} & 2.53 & 1.44 & .052 & \best{.9993} \\
Single shared RVQ & 640 & \best{64} & 3.18 & \best{1.34} & \best{.040} & \second{.9991} \\
No contact stream & 632 & \second{65} & \best{2.45} & 1.43 & .049 & .9962 \\
5 levels $\times$ 256 codes & 638 & 83 & \second{2.49} & 1.38 & .046 & \second{.9991} \\
8 levels $\times$ 32 codes & 638 & 137 & 2.73 & \second{1.37} & \second{.043} & .9990 \\
\bottomrule
\end{tabular}}
\end{table}

\subsection{Language: flow-matching policy}
\label{app:language}

\paragraph{Architecture.}
We use the first 16 language layers of the frozen Qwen3-VL-2B-Instruct backbone~\cite{bai2025qwen3vl} and a four-layer text refiner. The 32-layer DiT~\cite{peebles2023dit} has width 1{,}536 and adaptive normalization conditioned on flow time~\cite{perez2018film,peebles2023dit}. Each group of four layers includes text and image cross-attention. The image pathway is inactive during language training. Four 71-dimensional state observations are expressed relative to the latest root pose.

\paragraph{Training.}
Training mixes complete caption segments with continuation windows conditioned on a committed motion prefix. We mirror motion and captions together, drop text with probability 0.1 for classifier-free guidance~\cite{ho2022cfg}, and drop state with probability 0.5. Inference uses 32 Euler steps with guidance 3 for text-to-motion and 1.5 for streaming.

\subsection{Language: transition-aware post-training}
\label{app:switch}

\paragraph{Switch recipe.}
For SwitchBench, \alvl denotes the single-instruction policy and \alvls its transition-trained counterpart. Post-training adds 30k updates with the same architecture (about 3.2 hours on eight GPUs). We sample complete segments, continuation windows and instruction-switch chains in proportions 0.35, 0.40 and 0.25. Chains cover three replanning steps around a caption boundary. During roll-in, the prefix comes from the policy's prediction under the previous instruction. Short continuation windows expose the model to instructions that last only one or two seconds.

\paragraph{Matched-step ablation.}
On both smaller training pools, adding switch chains improves periodic rule success by over five percentage points relative to continuing the original objective for the same number of updates. The full recipe also includes roll-in and short continuations (\cref{tab:components}).

\begin{table}[t]
\centering
\caption{\textbf{SwitchBench training ablation.} Each post-training run uses 30k updates. Rankings are within each training pool. P and I denote periodic and interrupt requests. A star marks a rule-success increase over the preceding row whose paired 95\% confidence interval excludes zero.}
\label{tab:components}
\footnotesize
\setlength{\tabcolsep}{2.8pt}
\fitwidth{%
\begin{tabular}{l cc cc cc}
\toprule
 & \multicolumn{2}{c}{Rule success$\uparrow$} & \multicolumn{2}{c}{R@1$\uparrow$} & \multicolumn{2}{c}{Residue$\downarrow$} \\
\cmidrule(lr){2-3}\cmidrule(lr){4-5}\cmidrule(lr){6-7}
Post-training (30k steps) & P & I & P & I & P & I \\
\midrule
\multicolumn{7}{@{}l}{\textit{Capture-only pool}} \\
None (base model) & 29.0 & 34.4 & 27.2 & 35.4 & \second{28.8} & 25.1 \\
A: original recipe & 29.3 & 33.9 & 29.0 & 35.0 & 29.7 & 23.1 \\
B: + switch chains & \second{35.5}$^*$ & \second{43.9}$^*$ & \second{32.2} & \second{42.8} & 29.7 & \second{20.7} \\
\rowcolor{alvOurs}Full switch recipe & \best{39.9}$^*$ & \best{49.7}$^*$ & \best{37.3} & \best{48.3} & \best{23.9} & \best{16.7} \\
\midrule
\multicolumn{7}{@{}l}{\textit{Four-source pool}} \\
None (base model) & 29.9 & 34.7 & 26.4 & 34.6 & 31.7 & 27.4 \\
A: original recipe & 30.9 & 37.1 & 28.6 & 36.3 & 31.1 & 25.4 \\
B: + switch chains & \second{36.2}$^*$ & \second{46.3}$^*$ & \second{34.1} & \second{46.0} & \second{30.0} & \second{19.8} \\
\rowcolor{alvOurs}Full switch recipe & \best{39.6} & \best{49.2} & \best{37.9} & \best{50.8} & \best{25.9} & \best{19.0} \\
\bottomrule
\end{tabular}}
\end{table}

\begin{table*}[t]
\centering
\caption{\textbf{Evaluation sets and metrics.} Unprefixed table and figure numbers refer to the main paper.}
\label{tab:benchmarks}
\footnotesize
\setlength{\tabcolsep}{4pt}
\renewcommand{\arraystretch}{1.0}
\begin{tabularx}{\textwidth}{@{}l l Z{4.6cm} Y Z{3.3cm}@{}}
\toprule
Stage & Benchmark & Size & Measures & Used in \\
\midrule
Action & TK-test & 1{,}696 held-out recordings (15.5\,h) & joint and keypoint error, root drift and yaw, contact F1, jerk ratio, dead codes & \hyperref[tab:action-a]{Tab.~\ref*{tab:action}(a)}; \cref{tab:ablation} \\
 & Codec round trip & Noitom, LAFAN1-G1 (1{,}024), OmniRetarget (327) & joint and keypoint error, retrieval, executed MPJPE & \hyperref[tab:action-b]{Tab.~\ref*{tab:action}(b)} \\
\midrule
Language & Noitom val / test & 1{,}024 eight-second segments per split & R@$k$, FID, jerk, foot sliding & \cref{tab:main}, \cref{tab:raw} \\
 & OMG test & 1{,}024 segments & R@$k$, FID (released evaluator) & \cref{tab:main} \\
 & ReactiveBFM test & 992 captioned 100STYLE pairs & retrieval, jerk and foot sliding & \cref{tab:raw} \\
 & Development suite & 256 instructions $\times$ 4 seeds & rule-based instruction scores & \cref{tab:raw} \\
 & Sealed suite & 1{,}024 held-out requests & rule-based instruction scores & \cref{tab:main} \\
\midrule
Execution & Tracker rollouts & Noitom, OMG and ReactiveBFM & success and fall rate under 100\,N (three trackers) and 300\,N (two trackers) & \cref{tab:main,tab:trackers}, \cref{tab:push} \\
\midrule
Streaming & SwitchBench & 253 episodes, 752 switches & rule success, retrieval and residue & \cref{tab:streambench}, \cref{tab:components,tab:streaming,tab:categories} \\
 & Natural boundaries & 643 direction changes, 576 same-state controls & new motion state reached, with text and with a null prompt & \cref{tab:boundaries} \\
\midrule
Vision & HumanoidArena & 7 tasks $\times$ 60 episodes (3 seeds $\times$ 20) & SR, AFR & \cref{tab:humanoidarena} \\
 & SIMPLE & 9 tasks $\times$ 4 levels $\times$ 10 episodes & SR (L0--L2), success count (all levels) & \cref{tab:simple} \\
 & Physical G1 & 4 tasks, trials per task vary & task completion; execution metrics & \cref{fig:vla}, \cref{tab:realtrials} \\
\bottomrule
\end{tabularx}
\end{table*}

\subsection{Vision: post-training on demonstrations}
\label{app:vision-model}

\paragraph{Adapting the language-stage weights.}
The vision policy inherits the DiT trunk, text refiner and conditioning layers. New projections map the physical G1's 48-dimensional state and 51-dimensional action, or HumanoidArena's 64-dimensional state and 40-dimensional action, to the trunk. Both configurations predict 20-step chunks in the target action space.

On HumanoidArena, the new projections and image pathway train for 1k steps with the inherited weights fixed. We then fine-tune the action model using a lower learning rate for the inherited weights and evaluate exponential-moving-average weights. Vision post-training takes approximately 1.7--5.3 hours per task on four GPUs.

On SIMPLE, each task has its own policy conditioned on the instruction, camera images and a single robot state. We post-train \method for 15k updates per task with a global batch size of 256. The random-initialization baseline trains its action model from scratch on the same task demonstrations (\cref{tab:simple}).

\section{Experiments}
\label{app:exp}

\subsection{Benchmarks and metrics}
\label{app:protocol}

\paragraph{Motion reconstruction.}
We measure joint-angle MAE, body-local keypoint error, root position and yaw drift, and foot-contact F1. The jerk ratio divides reconstructed by reference jerk. Executed MPJPE measures joint position error after tracking the decoded motion.

\paragraph{Language--motion alignment.}
All text-to-motion baselines receive the benchmark caption and duration, and their outputs are converted to 50\,Hz G1 motion. R@$k$ measures paired-caption retrieval in batches of 32. FID~\cite{heusel2017fid} compares generated and reference motion embeddings. Noitom uses our T5-3B evaluator~\cite{raffel2020t5}. OMG and ReactiveBFM use the released OMG evaluator~\cite{huang2026omg}.

\paragraph{Fine-grained instruction following.}
The primary score averages gesture, laterality and order success. Locomotion is scored separately. Rules on the generated G1 kinematics check that the requested body part moves in the requested direction and that multi-step requests follow the requested order.

\paragraph{Physical quality and execution.}
Foot sliding measures foot speed during contact. Under the ReactiveBFM protocol~\cite{chen2026reactivebfm}, success requires no fall, pelvis-relative MPJPE below 100\,mm and planar root error below 3\,m at every frame. We count a fall when pelvis or torso height drops to $0.30$\,m or below while its reference remains above $0.30$\,m. \Cref{app:exp-push} specifies the disturbance schedule.

\paragraph{Following changing instructions.}
SwitchBench holds each command for 4--5\,s. Periodic requests enter at the next scheduled replan. Interrupts trigger an immediate replan. We score adoption of the new command and persistence of the old one 0.56--4\,s after each switch, using rules on root displacement, yaw and wrist position. The response time uses the first 0.3\,s of sustained positive new-minus-old caption similarity.

\paragraph{Vision.}
HumanoidArena reports success and fall rates~\cite{wang2026humanoidarena}. \Cref{tab:simple} reports SIMPLE success over L0--L2 and all four levels. On the physical G1, onboard video determines task completion and session logs supply the execution metrics (\cref{tab:realtrials}).

\onecolumn
\noindent\begin{minipage}[t]{.435\linewidth}\vspace{0pt}
\subsection{Effect of decoder look-ahead}
\label{app:exp-action}

\Cref{tab:lookahead} varies the number of future rows available during decoding. Relative to zero look-ahead, three rows reduce joint MAE from 0.0737 to 0.0110\,rad and raise retrieval R@1 from 60.8 to 81.1. This decoder adds 0.85\,mm of executed MPJPE relative to tracking the reference and uses 320\,ms less look-ahead than the seven-row decoder.
\end{minipage}\hfill
\begin{minipage}[t]{.535\linewidth}\vspace{0pt}
\par\noindent\begin{minipage}{\linewidth}
\captionsetup{type=table}
\centering
\caption{\textbf{Decoder look-ahead on Noitom validation.} Ripple is jerk energy at the 12.5\,Hz row frequency relative to the reference. Ripple is ranked by distance from 1. Extra MPJPE is the increase over tracking reference motion.}
\label{tab:lookahead}
\footnotesize
\setlength{\tabcolsep}{3.2pt}
\fitwidth{%
\begin{tabular}{l c cc c c c}
\toprule
 & Look-ahead & Joint & & & Limit & Extra MPJPE \\
Decoder & (ms) & MAE (rad)$\downarrow$ & Ripple & R@1$\uparrow$ & viol.\ (\%)$\downarrow$ & (mm)$\downarrow$ \\
\midrule
Ground truth & -- & 0 & \textcolor{alvGrey}{1} & \textcolor{alvGrey}{81.0} & \textcolor{alvGrey}{1.0} & -- \\
\midrule
Causal, $\ell{=}0$ & 0 & .0737 & 10.12 & 60.8 & 43.0 & 12.86 \\
Causal, $\ell{=}1$ & 80 & .0363 & 4.27 & -- & -- & -- \\
\rowcolor{alvOurs}Causal, $\ell{=}3$ (ours) & 240 & \second{.0110} & \second{1.85} & \best{81.1} & \second{1.4} & \second{0.85} \\
Context-aware, $\ell{=}7$ & 560 & \best{.0046} & \best{0.77} & \second{81.0} & \best{0.4} & \best{0.12} \\
\bottomrule
\end{tabular}}
\end{minipage}\par
\end{minipage}\par
\medskip
\noindent\begin{minipage}[t]{.435\linewidth}\vspace{0pt}
\subsection{Raw and learned action spaces}
\label{app:exp-raw}

We compare the learned action space with a policy that directly predicts four 71-dimensional state frames per row using the same training data and schedule. On the full language pool, development-suite instruction scores differ by 0.8 points. Across Noitom, OMG and ReactiveBFM, raw prediction has 53--87 times the jerk and 11--21 times the foot sliding of the learned model. On the smaller four-source pool, the learned action space also improves the primary instruction score by 7.7 points (\cref{tab:raw}).
\end{minipage}\hfill
\begin{minipage}[t]{.535\linewidth}\vspace{0pt}
\par\noindent\begin{minipage}{\linewidth}
\captionsetup{type=table}
\centering
\caption{\textbf{Raw and learned action spaces.} The primary score uses the development suite. Retrieval is R@3 on Noitom validation and ReactiveBFM. Jerk and foot sliding (m/s) use Noitom validation, with jerk ranked by distance from the reference value of 50.8. Rankings are within each training pool. A star marks a paired 95\% confidence interval excluding zero for the difference in primary score.}
\label{tab:raw}
\footnotesize
\setlength{\tabcolsep}{3.4pt}
\fitwidth{%
\begin{tabular}{l c cc cc}
\toprule
 & & \multicolumn{2}{c}{R@3$\uparrow$} & & \\
\cmidrule(lr){3-4}
Action space & Prim.$\uparrow$ & Noitom & RBFM & Jerk & Sliding$\downarrow$ \\
\midrule
\multicolumn{6}{@{}l}{\textit{Final pool}} \\
Raw motion & \best{64.2} & \best{91.4} & \best{62.1} & \second{2{,}828} & \second{.404} \\
\rowcolor{alvOurs}\method\ (learned) & \second{63.4} & \second{89.6} & \second{61.7} & \best{49.8} & \best{.023} \\
\midrule
\multicolumn{6}{@{}l}{\textit{Four-source pool}} \\
Raw motion & \second{54.7} & \best{84.9} & \best{47.6} & \second{2{,}925} & \second{.418} \\
\rowcolor{alvOurs}\method\ (learned) & \best{62.4}$^*$ & \second{83.4} & \second{47.3} & \best{42.7} & \best{.023} \\
\bottomrule
\end{tabular}}
\end{minipage}\par
\end{minipage}\par
\bigskip
\subsection{Execution under strong pushes}
\label{app:exp-push}

A prolonged rollout extends the same instruction to five times its nominal duration, capped at 32\,s. Each nominal interval receives a 0.1\,s horizontal push at a random angle and a start time sampled uniformly from its middle 60\%. The force is split equally between pelvis and torso. All methods use the same push schedule. Closed-loop planners update from the executed state.

Under 300\,N pushes, \method completes at least 95.6\% of episodes across the two trackers and three test sets in \cref{tab:push}. The largest baseline losses occur on the prolonged ReactiveBFM rollouts. With Humanoid-GPT, open-loop success there ranges from 7.4\% to 79.2\% for the three baselines, compared with 95.6\% for \method.

\smallskip
\par\noindent\begin{minipage}{\linewidth}
\captionsetup{type=table}
\centering
\caption{\textbf{Execution under 300\,N pushes.} Success rates (\%) at nominal ($1\times$) and prolonged ($5\times$) durations. The upper block uses open-loop execution and the lower block uses closed-loop replanning. Rankings are within each execution mode. RBFM denotes ReactiveBFM.}
\label{tab:push}
\footnotesize
\setlength{\tabcolsep}{4.2pt}
\fitwidth{%
\begin{tabular}{l ccc ccc ccc ccc}
\toprule
 & \multicolumn{6}{c}{Humanoid-GPT~\cite{qi2026humanoidgpt}} & \multicolumn{6}{c}{SONIC~\cite{luo2026sonic}} \\
\cmidrule(lr){2-7}\cmidrule(lr){8-13}
 & \multicolumn{3}{c}{1$\times$} & \multicolumn{3}{c}{5$\times$} & \multicolumn{3}{c}{1$\times$} & \multicolumn{3}{c}{5$\times$} \\
\cmidrule(lr){2-4}\cmidrule(lr){5-7}\cmidrule(lr){8-10}\cmidrule(lr){11-13}
Model & Noitom & OMG & RBFM & Noitom & OMG & RBFM & Noitom & OMG & RBFM & Noitom & OMG & RBFM \\
\midrule
Ground truth & \textcolor{alvGrey}{100.0} & \textcolor{alvGrey}{90.4} & \textcolor{alvGrey}{90.7} & -- & -- & -- & \textcolor{alvGrey}{99.8} & \textcolor{alvGrey}{93.4} & \textcolor{alvGrey}{99.7} & -- & -- & -- \\
\midrule
OMG-500M~\cite{huang2026omg} & \second{97.9} & \second{95.0} & \second{94.9} & 74.8 & 72.7 & 55.4 & 99.3 & \second{97.9} & \second{98.3} & 78.3 & 83.7 & 72.5 \\
Kimodo-G1~\cite{rempe2026kimodo} & 93.4 & 87.6 & 85.2 & 45.8 & \second{85.2} & 7.4 & 97.8 & 91.8 & 94.9 & 49.2 & \second{90.9} & 9.2 \\
ARDY-G1~\cite{zhao2026ardy} & 96.8 & 91.5 & 88.8 & \second{85.1} & 79.5 & \second{79.2} & \second{99.4} & 94.7 & 95.7 & \second{95.2} & 89.0 & \second{90.6} \\
\rowcolor{alvOurs}\method & \best{99.9} & \best{99.7} & \best{98.6} & \best{99.6} & \best{97.8} & \best{95.6} & \best{99.9} & \best{99.8} & \best{99.4} & \best{99.5} & \best{98.4} & \best{96.8} \\
\midrule
OMG-500M, closed loop & -- & -- & -- & \second{70.3} & \second{24.1} & \second{39.0} & -- & -- & -- & \second{69.8} & \second{25.2} & \second{40.6} \\
\rowcolor{alvOurs}\method, closed loop & -- & -- & -- & \best{99.5} & \best{97.9} & \best{95.7} & -- & -- & -- & \best{99.0} & \best{98.4} & \best{96.3} \\
\bottomrule
\end{tabular}}
\end{minipage}\par

\clearpage
\subsection{SwitchBench and natural transitions}
\label{app:exp-streaming}

\paragraph{Effect across training pools.}
Transition training improves rule success on all three pools by 9.7--13.4 points under periodic requests and 12.4--14.5 points under interrupts (\cref{tab:streaming}). Old-command residue also falls under both schedules.

\smallskip
\par\noindent\begin{minipage}{\linewidth}
\captionsetup{type=table}
\centering
\caption{\textbf{SwitchBench by training pool.} Rankings compare the two models trained on the same data. P and I denote periodic and interrupt requests. A star marks a paired change from the single-instruction model whose 95\% confidence interval excludes zero.}
\label{tab:streaming}
\footnotesize
\setlength{\tabcolsep}{5pt}
\begin{tabular}{l l cc cc c cc}
\toprule
 & & \multicolumn{2}{c}{Rule success$\uparrow$} & \multicolumn{2}{c}{R@1$\uparrow$} & R@3$\uparrow$ & \multicolumn{2}{c}{Residue$\downarrow$} \\
\cmidrule(lr){3-4}\cmidrule(lr){5-6}\cmidrule(lr){7-7}\cmidrule(lr){8-9}
Model & Language data & P & I & P & I & P & P & I \\
\midrule
\alvl & Four-source & \second{29.9} & \second{34.7} & \second{26.4} & \second{34.6} & \second{39.8} & \second{31.7} & \second{27.4} \\
\quad \alvls & Four-source & \best{39.6}$^*$ & \best{49.2}$^*$ & \best{37.9}$^*$ & \best{50.8}$^*$ & \best{52.7}$^*$ & \best{25.9}$^*$ & \best{19.0}$^*$ \\
\alvl & OMG-extended & \second{27.7} & \second{38.6} & \second{30.4} & \second{37.6} & \second{43.9} & \second{35.6} & \second{25.0} \\
\quad \alvls & OMG-extended & \best{39.4}$^*$ & \best{50.9}$^*$ & \best{36.9}$^*$ & \best{49.5}$^*$ & \best{52.1}$^*$ & \best{22.5}$^*$ & \best{14.9}$^*$ \\
\rowcolor{alvOurs}\alvl & Final & \second{26.6} & \second{35.8} & \second{27.2} & \second{35.9} & \second{40.4} & \second{34.4} & \second{24.8} \\
\rowcolor{alvOurs}\quad \alvls & Final & \best{40.0}$^*$ & \best{48.8}$^*$ & \best{37.3}$^*$ & \best{47.7}$^*$ & \best{53.6}$^*$ & \best{26.6}$^*$ & \best{18.0}$^*$ \\
\bottomrule
\end{tabular}
\end{minipage}\par

\paragraph{Which transitions improve?}
Under periodic requests, the largest gain by target category is for backward walking (24.0 points). With interrupts it rises to 38.5 points (\cref{tab:categories}). TextOp has the highest periodic success for forward walking and ARDY-G1 for stopping. \Cref{fig:supp-qual-switch} shows two transitions from matched instructions.

\smallskip
\par\noindent\begin{minipage}{\linewidth}
\captionsetup{type=table}
\centering
\caption{\textbf{Rule success by switch type.} Rows group switches by the new instruction (top) and preceding motion (bottom). $\Delta$ gives the difference between \alvls and \alvl in percentage points. A star marks a paired 95\% confidence interval excluding zero. Rankings compare models within each row and request schedule.}
\label{tab:categories}
\footnotesize
\setlength{\tabcolsep}{4.4pt}
\begin{tabular}{l c cc *{2}{>{\columncolor{alvOurs}}c} c *{2}{>{\columncolor{alvOurs}}c} c}
\toprule
 & & \multicolumn{5}{c}{Periodic requests} & \multicolumn{3}{c}{Interrupt requests} \\
\cmidrule(lr){3-7}\cmidrule(lr){8-10}
Switch & $n$ & TextOp~\cite{xie2026textop} & ARDY-G1~\cite{zhao2026ardy} & \alvl & \alvls & $\Delta$ & \alvl & \alvls & $\Delta$ \\
\midrule
to standing & 96 & 9.4 & \best{55.2} & 13.5 & \second{20.8} & +7.3 & \second{21.9} & \best{30.2} & +8.3 \\
to walking forward & 112 & \best{70.5} & 18.8 & 36.6 & \second{41.1} & +4.5 & \second{40.2} & \best{51.8} & +11.6$^*$ \\
to walking backward & 96 & 4.2 & \second{16.7} & 12.5 & \best{36.5} & +24.0$^*$ & \second{16.7} & \best{55.2} & +38.5$^*$ \\
to sidestepping & 112 & 0.9 & \second{18.8} & 12.5 & \best{24.1} & +11.6$^*$ & \second{19.6} & \best{33.9} & +14.3$^*$ \\
to turning & 112 & 29.5 & 31.2 & \second{49.1} & \best{68.8} & +19.6$^*$ & \second{69.6} & \best{79.5} & +9.8 \\
to arm gesture & 112 & \second{36.6} & 21.4 & 34.8 & \best{53.6} & +18.8$^*$ & \second{52.7} & \best{56.2} & +3.6 \\
to walking with gesture & 112 & 11.6 & 17.0 & \second{23.2} & \best{32.1} & +8.9 & \second{25.0} & \best{33.0} & +8.0 \\
\midrule
from standing & 96 & 27.1 & \second{32.3} & 21.9 & \best{46.9} & +25.0$^*$ & \second{41.7} & \best{60.4} & +18.8$^*$ \\
from moving & 544 & 25.2 & 25.9 & \second{27.9} & \best{39.0} & +11.0$^*$ & \second{35.8} & \best{47.6} & +11.8$^*$ \\
from arm gesture in place & 112 & 15.2 & 15.2 & \second{24.1} & \best{39.3} & +15.2$^*$ & \second{30.4} & \best{44.6} & +14.3 \\
\bottomrule
\end{tabular}
\end{minipage}\par

\par\medskip
\noindent\begin{minipage}[c]{.40\linewidth}
\paragraph{Time to adopt the new instruction.}
For switches initially aligned with the old instruction, \alvl reaches positive alignment with the new instruction by 2\,s on 26.6\% of interrupt requests and 11.8\% of periodic requests. Under periodic requests, 82.7\% of \alvls transitions have turned by 4\,s, compared with 60.8\% for \alvl (\cref{fig:supp-dynamics}).
\end{minipage}\hfill
\begin{minipage}[c]{.56\linewidth}
\par\noindent\begin{minipage}{\linewidth}
\captionsetup{type=figure}
\centering
\includegraphics[width=\linewidth]{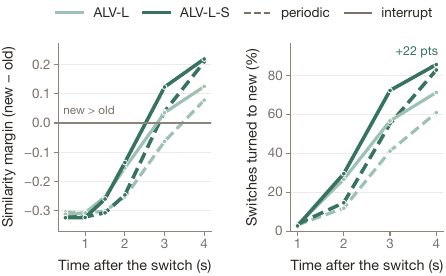}
\caption{\textbf{Response after an instruction change.} The left panel shows the mean similarity margin for switches initially aligned with the old caption (526 periodic, 546 interrupt). The right panel shows the fraction with a positive margin sustained for 0.3\,s by time $t$.}
\label{fig:supp-dynamics}
\end{minipage}\par
\end{minipage}\par
\clearpage
\paragraph{Planning window and inference time.}
At 80\,ms per latent row, both ALV variants predict $W=40$ rows, replan every $P=7$ rows in periodic mode, and retain $C=12$ committed rows. Interrupts trigger an immediate replan with $d+\ell=10$ rows fixed (planning delay $d=7$, decoder look-ahead $\ell=3$). On an NVIDIA RTX~4090, 32 Euler steps with batched guidance took 442\,ms at the 95th percentile. With 32.5\,ms for decoding and 20\,ms for transfer, the delay budget rounds up to seven rows. ALV's SwitchBench rollouts ran on an NVIDIA H200.

\paragraph{Planner inputs and response latency.}
All planners shared the instruction sequences, Humanoid-GPT tracker and 0.56\,s delay budget. ALV read the robot state at every replan. ARDY-G1 and TextOp used their own generated motion histories and updated at their native step boundaries after the delay. Across 752 switches, the interval to the first reference frame the new instruction could affect was 1.22\,s for periodic ALV, 0.80\,s for interrupt ALV, 0.56\,s for ARDY-G1 and 0.60\,s for TextOp on the simulation clock. \Cref{fig:supp-dynamics} gives the response measured in executed motion. Physical G1 onset latency is unmeasured in this benchmark.

\par\medskip
\par\noindent\begin{minipage}{\linewidth}
\captionsetup{type=figure}
\centering
\includegraphics[width=.69\textwidth]{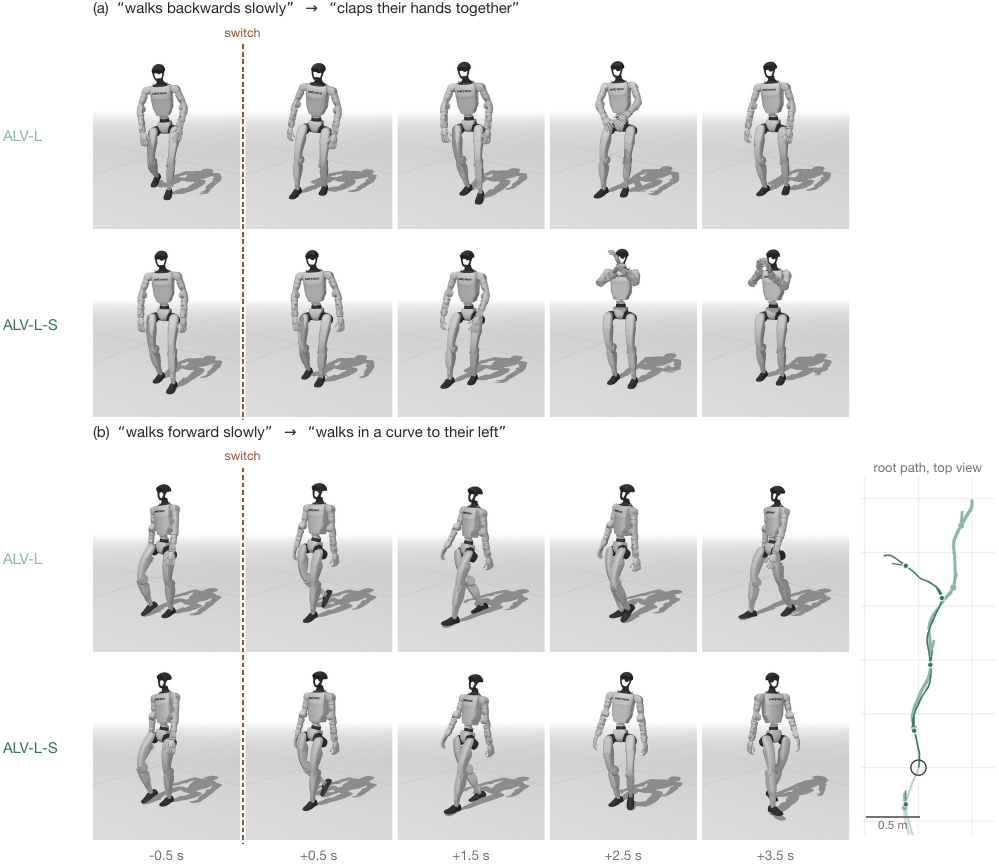}
\caption{\textbf{Executed motion after a SwitchBench instruction change.} The circle marks the switch and the paths show root motion under periodic requests. In (a), \alvls stops and claps while \alvl continues walking backward. In (b), \alvls turns left and \alvl continues straight.}
\label{fig:supp-qual-switch}
\end{minipage}\par

\par\medskip\noindent\begin{minipage}[t]{.32\linewidth}\vspace{0pt}
\paragraph{Natural caption boundaries.}
We condition both policies on the same recorded prefix at a caption boundary and score whether the continuation reaches the annotated motion state (\cref{tab:boundaries}). With the instruction omitted, transition training raises direction-change success from 53.2\% to 62.1\% on the full pool. On same-state controls, both models preserve the original state in over 96\% of cases.
\end{minipage}\hfill
\begin{minipage}[t]{.65\linewidth}\vspace{0pt}
\captionsetup{type=table}
\centering
\caption{\textbf{Transitions at natural caption boundaries.} Entries give the percentage of continuations that reach the annotated state after a recorded prefix. Text and null use the new instruction and an empty prompt, respectively. Rankings compare the two models trained on the same data. A star marks a paired improvement whose 95\% confidence interval excludes zero.}
\label{tab:boundaries}
\footnotesize
\setlength{\tabcolsep}{4pt}
\begin{tabular}{l cc cc}
\toprule
 & \multicolumn{2}{c}{Direction change (643)$\uparrow$} & \multicolumn{2}{c}{Same state (576)} \\
\cmidrule(lr){2-3}\cmidrule(lr){4-5}
Model & Text & Null & Text & Null \\
\midrule
\alvl (four-source) & \second{72.5} & \second{49.0} & \best{98.4} & \best{96.9} \\
\alvls (four-source) & \best{79.8}$^*$ & \best{62.5}$^*$ & \second{97.9} & \second{96.4} \\
\alvl (OMG-ext.) & \second{72.5} & \second{50.2} & \best{97.4} & \best{97.6} \\
\alvls (OMG-ext.) & \best{79.9}$^*$ & \best{61.0}$^*$ & \second{96.7} & \second{96.0} \\
\rowcolor{alvOurs}\alvl & \second{75.7} & \second{53.2} & \best{97.4} & \best{97.4} \\
\rowcolor{alvOurs}\alvls & \best{79.3}$^*$ & \best{62.1}$^*$ & \second{96.7} & \second{96.4} \\
\bottomrule
\end{tabular}
\end{minipage}\par

\clearpage
\enlargethispage{5pt}
\subsection{Vision: real-robot post-training}
\label{app:exp-real}

We post-train \method and the GR00T N1.7 action head on the same teleoperated demonstrations for each household task. Using head and wrist images, proprioception and a task instruction, both policies completed each task on the physical G1 (\cref{tab:realtrials}). For each trial, \cref{tab:realtrials} reports its execution metrics. Per-trial median inference times are 69--85\,ms for both policies, and the pelvis stays above 0.74\,m in every trial.

\par\smallskip\noindent\begin{minipage}{\linewidth}
\centering
\captionsetup{type=table,font=footnotesize}
\caption{\textbf{Physical G1 task completion and execution metrics of the trials.} One row per trial. Each mean row gives successful completions among the five trials and averages the metrics; bold and underline rank the two counts within each task. Metrics are computed from the robot logs, over the ticks controlled by the policy: task time; RMS jerk of the 29 body joints, of the executed end effectors (both wrists and ankles) and of the commanded end effectors; mean end-effector jump at chunk replacements; chunk roughness of the planned wrist keyframes; mean error of the executed end effectors and pelvis against the command; mean right-hand error between command and measured BrainCo positions ($\times 10^{-3}$); maximum torso tilt; median policy inference time. Lower is smoother or closer for every metric except time.}
\label{tab:realtrials}
\footnotesize
\setlength{\tabcolsep}{3.6pt}
\renewcommand{\arraystretch}{.84}
\fitwidth{%
\begin{tabular}{@{}l c c c cc c c cc c c c@{}}
\toprule
 & & Time & Joint jerk & \multicolumn{2}{c}{EE jerk (m/s$^3$)} & Swap jump & Roughness & \multicolumn{2}{c}{Tracking (mm)} & Grip & Tilt & Inference \\
\cmidrule(lr){5-6}\cmidrule(lr){9-10}
Policy & Result & (s) & (rad/s$^3$) & executed & commanded & (mm) & wrist (mm) & EE & pelvis & error & ($^\circ$) & (ms) \\
\midrule
\multicolumn{13}{@{}l}{\textit{Throw rubbish into the box}} \\
\method\ (ours) & \checkmark & 32.0 & 364 & 32.8 & 58.4 & 4.1 & 14.1 & 37.0 & 39.2 & 4.2 & 10.2 & 70 \\
 & \checkmark & 18.7 & 451 & 49.5 & 60.5 & 4.3 & 13.2 & 43.4 & 38.5 & 5.8 & 9.6 & 70 \\
 & \checkmark & 18.8 & 452 & 39.1 & 61.9 & 4.9 & 14.1 & 39.3 & 37.9 & 7.1 & 7.5 & 69 \\
 & \checkmark & 28.2 & 337 & 34.7 & 60.3 & 4.5 & 14.0 & 44.8 & 44.7 & 11.4 & 9.5 & 70 \\
 & $\times$ & 23.6 & 396 & 38.5 & 59.5 & 4.5 & 14.0 & 42.1 & 36.0 & 6.2 & 9.0 & 70 \\
\rowcolor{alvOurs}\quad\textit{mean} & \best{4/5} & 24.3 & 400 & 38.9 & 60.1 & 4.5 & 13.9 & 41.3 & 39.2 & 7.0 & 9.2 & 70 \\
\cmidrule(lr){1-13}
GR00T N1.7 head & \checkmark & 17.8 & 404 & 46.0 & 55.6 & 4.8 & 13.0 & 41.3 & 35.3 & 5.2 & 8.1 & 70 \\
 & \checkmark & 21.4 & 331 & 31.9 & 55.3 & 4.4 & 12.7 & 30.7 & 31.2 & 6.4 & 7.9 & 69 \\
 & $\times$ & 19.0 & 432 & 37.2 & 54.7 & 4.2 & 13.0 & 36.5 & 30.4 & 3.2 & 12.3 & 70 \\
 & \checkmark & 19.1 & 362 & 31.8 & 55.7 & 4.8 & 13.0 & 30.4 & 29.8 & 7.4 & 7.8 & 70 \\
 & $\times$ & 26.5 & 272 & 23.6 & 53.8 & 3.4 & 12.6 & 29.4 & 32.2 & 6.4 & 7.5 & 70 \\
\quad\textit{mean} & \second{3/5} & 20.8 & 360 & 34.1 & 55.0 & 4.3 & 12.8 & 33.6 & 31.8 & 5.7 & 8.7 & 70 \\
\midrule
\multicolumn{13}{@{}l}{\textit{Pull the chair}} \\
\method\ (ours) & \checkmark & 36.8 & 162 & 14.2 & 62.0 & 3.7 & 15.8 & 57.1 & 29.3 & 29.6 & 7.0 & 70 \\
 & \checkmark & 24.9 & 196 & 17.2 & 60.5 & 3.4 & 14.9 & 53.5 & 31.8 & 8.9 & 6.6 & 70 \\
 & \checkmark & 17.5 & 420 & 39.0 & 59.6 & 3.4 & 14.8 & 47.3 & 29.8 & 15.5 & 9.9 & 70 \\
 & \checkmark & 15.0 & 254 & 26.9 & 63.4 & 3.3 & 15.5 & 57.1 & 43.7 & 6.9 & 10.5 & 70 \\
 & $\times$ & 8.4 & 260 & 42.1 & 63.3 & 3.2 & 14.7 & 46.1 & 39.6 & 6.4 & 13.9 & 70 \\
\rowcolor{alvOurs}\quad\textit{mean} & \best{4/5} & 20.5 & 259 & 27.9 & 61.8 & 3.4 & 15.2 & 52.2 & 34.8 & 13.5 & 9.6 & 70 \\
\cmidrule(lr){1-13}
GR00T N1.7 head & \checkmark & 14.3 & 206 & 18.7 & 75.4 & 4.5 & 18.1 & 40.5 & 28.6 & 13.5 & 6.6 & 70 \\
 & $\times$ & 15.2 & 191 & 16.6 & 75.2 & 4.1 & 18.2 & 28.5 & 25.1 & 6.5 & 5.7 & 70 \\
 & $\times$ & 13.6 & 217 & 18.4 & 73.1 & 3.6 & 17.4 & 26.5 & 20.1 & 7.8 & 4.8 & 70 \\
 & \checkmark & 13.3 & 298 & 26.7 & 77.3 & 4.6 & 18.7 & 35.1 & 20.4 & 11.6 & 6.8 & 70 \\
 & $\times$ & 12.1 & 302 & 27.1 & 79.7 & 5.3 & 18.7 & 50.8 & 40.3 & 6.3 & 6.7 & 70 \\
\quad\textit{mean} & \second{2/5} & 13.7 & 243 & 21.5 & 76.2 & 4.4 & 18.2 & 36.3 & 26.9 & 9.1 & 6.1 & 70 \\
\midrule
\multicolumn{13}{@{}l}{\textit{Pull the curtain}} \\
\method\ (ours) & \checkmark & 13.3 & 395 & 42.3 & 91.8 & 6.4 & 24.4 & 45.4 & 48.6 & 8.8 & 6.4 & 69 \\
 & \checkmark & 13.7 & 367 & 39.6 & 89.0 & 6.0 & 24.6 & 43.8 & 42.6 & 8.6 & 9.2 & 69 \\
 & \checkmark & 13.9 & 366 & 39.9 & 90.7 & 6.4 & 25.2 & 44.6 & 42.7 & 8.9 & 6.4 & 70 \\
 & \checkmark & 16.2 & 377 & 40.1 & 92.2 & 6.7 & 24.9 & 41.6 & 41.5 & 10.3 & 7.1 & 73 \\
 & \checkmark & 12.3 & 359 & 34.2 & 90.5 & 7.4 & 25.4 & 41.2 & 36.7 & 8.7 & 6.9 & 85 \\
\rowcolor{alvOurs}\quad\textit{mean} & \best{5/5} & 13.9 & 373 & 39.2 & 90.9 & 6.6 & 24.9 & 43.3 & 42.4 & 9.1 & 7.2 & 73 \\
\cmidrule(lr){1-13}
GR00T N1.7 head & \checkmark & 12.3 & 454 & 47.7 & 85.8 & 6.4 & 22.6 & 58.2 & 59.8 & 11.5 & 7.2 & 69 \\
 & \checkmark & 11.3 & 417 & 39.2 & 85.4 & 6.4 & 23.3 & 44.1 & 42.6 & 11.8 & 10.1 & 69 \\
 & $\times$ & 11.5 & 342 & 31.2 & 81.3 & 5.7 & 21.2 & 36.4 & 41.5 & 12.5 & 9.3 & 70 \\
 & \checkmark & 10.5 & 370 & 36.1 & 82.5 & 6.4 & 22.5 & 47.5 & 42.2 & 9.8 & 9.6 & 70 \\
 & \checkmark & 11.8 & 384 & 37.6 & 81.8 & 6.2 & 21.6 & 43.0 & 42.0 & 9.6 & 9.5 & 70 \\
\quad\textit{mean} & \second{4/5} & 11.5 & 393 & 38.3 & 83.4 & 6.2 & 22.2 & 45.8 & 45.6 & 11.0 & 9.1 & 70 \\
\midrule
\multicolumn{13}{@{}l}{\textit{Retrieve clothes from the wardrobe}} \\
\method\ (ours) & \checkmark & 35.3 & 213 & 20.0 & 59.6 & 3.7 & 15.1 & 44.5 & 31.3 & 5.0 & 7.8 & 70 \\
 & \checkmark & 25.9 & 274 & 30.0 & 60.9 & 4.4 & 15.8 & 40.7 & 38.7 & 10.5 & 6.7 & 70 \\
 & $\times$ & 32.9 & 175 & 13.7 & 55.0 & 3.4 & 14.5 & 36.4 & 19.5 & 4.5 & 4.4 & 70 \\
 & $\times$ & 36.2 & 167 & 13.5 & 55.2 & 3.0 & 14.7 & 37.8 & 16.0 & 5.7 & 5.5 & 70 \\
 & $\times$ & 13.9 & 318 & 36.3 & 62.7 & 3.7 & 15.6 & 42.9 & 34.4 & 8.0 & 11.3 & 70 \\
\rowcolor{alvOurs}\quad\textit{mean} & \best{2/5} & 28.8 & 230 & 22.7 & 58.7 & 3.6 & 15.2 & 40.5 & 28.0 & 6.7 & 7.2 & 70 \\
\cmidrule(lr){1-13}
GR00T N1.7 head & $\times$ & 15.0 & 247 & 20.2 & 59.2 & 4.0 & 15.6 & 40.3 & 35.1 & 9.3 & 7.9 & 70 \\
 & $\times$ & 8.6 & 372 & 32.4 & 63.8 & 5.4 & 15.4 & 35.1 & 42.0 & 5.6 & 11.5 & 70 \\
 & $\times$ & 14.6 & 333 & 28.6 & 65.7 & 4.4 & 16.0 & 42.1 & 40.7 & 6.4 & 11.3 & 70 \\
 & \checkmark & 23.9 & 275 & 30.4 & 66.6 & 4.4 & 16.1 & 41.4 & 60.9 & 4.9 & 7.7 & 70 \\
 & $\times$ & 11.3 & 278 & 23.8 & 59.0 & 4.4 & 15.6 & 39.7 & 39.7 & 6.3 & 8.7 & 70 \\
\quad\textit{mean} & \second{1/5} & 14.7 & 301 & 27.1 & 62.9 & 4.5 & 15.8 & 39.7 & 43.7 & 6.5 & 9.4 & 70 \\
\bottomrule
\end{tabular}}
\end{minipage}\par

\end{document}